\documentclass[5p,times]{elsarticle}

\usepackage{lineno,hyperref}
\modulolinenumbers[5]

\usepackage{tikz}

\usepackage{graphicx}
\usepackage{amsmath}
\usepackage{amssymb}

\usepackage[toc,page]{appendix}

\definecolor{mygreen}{RGB}{0, 190, 0}

\usepackage[scientific-notation=true]{siunitx}

\usepackage{hyperref} % For hyperlinks in the PDF

\usepackage[normalem]{ulem}

\usepackage{arydshln}

\usepackage{natbib}

\journal{Journal of \LaTeX\ Templates}

\biboptions{authoryear}

\hypersetup{draft}
\begin{document}

\sloppy

\begin{frontmatter}

\title{Weakly Supervised Object Detection with 2D and 3D Regression Neural Networks}

\author[bigr]{Florian~Dubost}
\cortext[mycorrespondingauthor]{Corresponding author}
\ead{floriandubost1@gmail.com}

\author[epi]{Hieab~Adams}
\author[epi]{Pinar~Yilmaz}
\author[bigr]{Gerda~Bortsova}
\author[bigr]{Gijs~van~Tulder}
\author[neuroepi]{M.~Arfan~Ikram}
\author[bigr,tudelft]{Wiro~Niessen}
\author[epi]{Meike~W.~Vernooij}
\author[bigr,dtu]{Marleen~de~Bruijne}
\ead{marleen.debruijne@erasmusmc.nl}

\address[bigr]{Biomedical Imaging Group Rotterdam, Departments of Radiology and Medical Informatics, Erasmus MC - University Medical Center Rotterdam, The Netherlands}
\address[epi]{Departments of Radiology and Nuclear Medicine, and Epidemiology, Erasmus MC - University Medical Center Rotterdam, The Netherlands}
\address[neuroepi]{Departments of Radiology, Epidemiology and Neurology. Erasmus MC - University Medical Center Rotterdam, The Netherlands}
\address[tudelft]{Department of Imaging Physics, Faculty of Applied Science, TU Delft, Delft, The Netherlands}
\address[dtu]{Machine Learning Section, Department of Computer Science, University of Copenhagen, Copenhagen, Denmark}

\begin{abstract}
Finding automatically multiple lesions in large images is a common problem in medical image analysis. Solving this problem can be challenging if, during optimization, the automated method cannot access information about the location of the lesions nor is given single examples of the lesions. We propose a new weakly supervised detection method using neural networks, that computes attention maps revealing the locations of brain lesions. These attention maps are computed using the last feature maps of a segmentation network optimized only with global image-level labels. The proposed method can generate attention maps at full input resolution without need for interpolation during preprocessing, which allows small lesions to appear in attention maps. For comparison, we modify state-of-the-art methods to compute attention maps for weakly supervised object detection, by using a global regression objective instead of the more conventional classification objective. This regression objective optimizes the number of occurrences of the target object in an image, e.g. the number of brain lesions in a scan, or the number of digits in an image. We study the behavior of the proposed method in MNIST-based detection datasets, and evaluate it for the challenging detection of enlarged perivascular spaces -- a type of brain lesion -- in a dataset of 2202 3D scans with point-wise annotations in the center of all lesions in four brain regions. In MNIST-based datasets, the proposed method outperforms the other methods. In the brain dataset, the weakly supervised detection methods come close to the human intrarater agreement in each region. The proposed method reaches the best area under the curve in two out of four regions, and has the lowest number of false positive detections in all regions, while its average sensitivity over all regions is similar to that of the other best methods. The proposed method can facilitate epidemiological and clinical studies of enlarged perivascular spaces and help advance research in the etiology of enlarged perivascular spaces and in their relationship with cerebrovascular diseases.

\end{abstract}


\end{frontmatter}

\section{Introduction}

Weakly supervised machine learning methods are designed to be optimized with limited amounts of labelled data and are very promising for a large number of medical image analysis problems. As medical expertise is scarce and annotation time expensive, unsupervised \citep{schlegl2017} and weakly supervised methods \citep{qi2017,bortsova2018} are most suited to extract information from large medical databases, in which labels are often either sparse or non-existent. In this article, we use attention maps for weakly supervised detection of brain lesions. Attention maps can be computed to reveal discriminative areas for the predictions of neural networks that process images such MRI, CT or X-ray. Most attention maps computation methods have originally been designed to make deep networks more explainable \citep{zhang2018,oktay2018,zhang2018b,hwang2016}. 
As those methods do not require annotations for the optimization of the networks but only global labels such as biomarkers or phenotypes \citep{wang2019}, they can also be optimized using only counting objectives such as the number of lesions in a brain region, and subsequently predict the location of these lesions during test time.

We propose a novel weakly supervised detection method, using attention maps computed from the feature maps of a segmentation network architecture optimized with global labels. By using the last feature maps of such an architecture, attention maps can be computed at full input resolution, and small structures can be detected more accurately. In this article, we focus on weak supervision with regression neural networks for counting. Regression networks have widely been optimized with local labels such as voxel coordinates \citep{redmon2016}, distance maps \citep{xie2018, xie2018b} or depth maps \citep{laina2016}.

Less frequently, regression networks have been used to predict global labels, such as age \citep{cole2017,wang2019}, brain lesion count \citep{dubost2017}, pedestrian count \citep{segui2015}, or car count \citep{mundhenk2016}. Other researchers have also optimized neural networks to infer count. \cite{ren2017} combined a recurrent network with an attention model to jointly count and segment the target objects, but need pixel-wise ground truths for the optimization. In bioimaging, methods inferring count have often been applied to cell counting in 2D images \citep{lempitsky2010, walach2016,xie2018,tan2018,alam2019}. These approaches are often optimized to regress distance or density maps computed from dot annotations at the center of the target objects. Instead of regressing density maps, \cite{paul2017} performed cell counting by regressing pixel-wise labels that represent the count of cells in the neighborhood. In our approach, pixel-wise labels are not needed for training: only the image-level count are used. Earlier, \cite{segui2015} have also optimized networks using image-level count labels alone for digit and pedestrian count and visualized the attention of the networks. However, they did not quantify the performance of the resulting weakly supervision detection. \cite{xue2016} performed cell counting also using regression network optimized with patch-wise cell count, computed density maps, but did not quantify the performance on the pixel level. In this article, we optimize regression networks using image-level count labels, but use this as a means for detection.

We compare the proposed method to four state-of-the-art methods \citep{simonyan2014,springenberg2014,schlemper2018,selvaraju2017}. Other weakly supervised detection methods have been proposed relying, for example, on latent support vector machines (SVMs) \citep{felzenszwalb2010}, a reformulation of the multiple instance learning mi-SVMs \citep{andrews2003}, or more recently, on multiple instance learning with attention-based neural networks \citep{ilse2018}, and on iterative learning with neural networks classifiers, where the training set is made of subsets of most reliable bounding boxes from the last iteration \cite{sangineto2018}.

We evaluate the methods using two datasets: a MNIST-based detection dataset and a dataset for the detection of enlarged perivascular spaces, a type of brain lesion that is associated with cerebral small vessel disease. On 1.5T scans, perivascular spaces become visible when enlarged. Following the neuroimaging standards proposed by \cite{wardlaw2013}, we use the consensus term perivascular space (PVS) throughout the manuscript without always referring to their enlargement. PVS is an emerging biomarker, and ongoing research attempts to better understand their etiology and relation with neurological disorders \citep{adams2014,duperron2019,gutierrez2019}. Most of the research on perivascular spaces is based on quantification of PVS burden using visual scores based PVS counts \citep{adams2014,potter2015}. Next to overall PVS burden, the location of PVS can have a clinical significance that varies depending on the brain region (midbrain, hippocampi, basal ganglia and centrum semiovale) and also within a brain region. For example PVS are thought to be benign when observed where perforating vessels enter the brain region \citep{jungreis1988}, such as PVS in the lower half of the basal ganglia. Understanding more precisely how the specific locations of PVS can relate with determinants of PVS and outcomes can aid neurology research. Automatically quantifying and detecting PVS is challenging, because PVS are very small (at the limit of the scan resolution) and can easily be confused with several other types of lesions \citep{dubost2018,Adams2013,sudre2018,brown2018}. Recently, automated methods have been developed to address PVS quantification \citep{ballerini2018,sudre2018,sepehrband2019,boespflug2018}, but these methods were not evaluated in large datasets or for the detection of individual PVS.
The proposed method only requires PVS visual scores for its optimization and is evaluated for the detection of individual PVS. In most of the large imaging studies, PVS are quantified using visual scores based on counts. Considering the generalizability issues of neural networks, using networks that require only PVS count for their optimization can consequently be considered to have more practical impact than networks that require annotations for their optimization.

 \subsection{State-of-the-art for attention map computation}
\label{sec:stateArtWeak}

All state-of-the-art methods investigated in this article are based on convolutional neural networks (CNNs) that compute a pseudo-probability map which indicates the locations of the target objects in the input image. In the rest of the article, we call this map the \textit{attention map}. The methods can be divided into three categories: methods using class activation maps (CAMs), methods based on the gradient of the output of the network, and methods using perturbations of the input of the network.

\paragraph{CAM methods}
This category consists of variants of the class activation maps (CAMs) method proposed by \cite{zhou2016}. CAMs are computed from the deepest feature maps of the network. These feature maps are followed by a global pooling layer, and usually one or more fully connected layers to connect to the output of the network. CAMs are computed during inference as a linear combination of these last feature maps, weighted by the parameters of the fully connected layers learnt during training. If the last feature maps have a much lower resolution than the input -- as is the case in deep networks with multiple pooling layers -- the resulting attention maps can be very coarse. This is suboptimal when small objects need to be localized, or when contours need to be segmented precisely. To alleviate this issue, \cite{dubost2017,schlemper2018} proposed to include finer-scale and lower-level feature maps in the computation of the attention maps. \cite{dubost2017} combined higher and lower level feature maps via skip connections and concatenation similarly to U-Net \citep{ronneberger2015}, while \cite{schlemper2018} used gated attention mechanisms, which rely on the implicit computation of internal attention maps. \cite{selvaraju2017} proposed to generalize CAM to any network architecture, using weights computed with the derivative of the output. Unlike other CAM methods, the method by \cite{selvaraju2017} does not require the presence of a global pooling layer in the network, and can be computed for any layer of the network.

\paragraph{Gradient methods}
\cite{simonyan2014} proposed to compute attention maps using the derivative of a classification network’s output with respect to the input image. These attention maps are fine-grained, but often noisy. \cite{springenberg2014} reduced this noise by  masking the values corresponding to negative entries of the top gradient (coming from the output of the network) in the ReLU activations. Gradients methods can be applied to any CNN.

\paragraph{Perturbation methods}
Perturbation methods compute attention maps by applying random perturbations to the input and observe the changes in the network output. These methods are model-agnostic, they can be used with any prediction model, not even necessarily restricted to neural networks. One of the simplest and most effective implementations of such methods was recently proposed by \cite{petsiuk2018} with masking perturbations. The input is masked with a series of random smooth masks, before being passed to the network. Using a linear combination of these masks weighted by the updated network classification scores, the authors could compute attention maps revealing the location of the target object. This method relies on a mask sampling technique, where the masks are first sampled in a lower dimensional space, and then rescaled to the size of the full image. Earlier, \cite{fong2017} proposed several other perturbation techniques including replacing a region with a constant value, injecting noise, and blurring the image.  Perturbation methods are the most general as they can also be applied to other classifiers than CNN. We do not study perturbation models in this paper, because their optimization was more challenging than that of other methods, especially for the detection of small objects.

\subsection{Contributions}

The contribution of this work is fourfold.
First, we propose a novel weakly-supervised detection method, named \textit{GP-Unet}. The principle of the method is to use a segmentation architecture with skip connections to compute attention maps at full input resolution to help the detection of small objects. A preliminary version of this work was presented in \citep{dubost2017}. 

Second, the proposed method is compared to five previously published methods \citep{dubost2017,schlemper2018,selvaraju2017,simonyan2014,springenberg2014}.

Third, we assess in MNIST-based \citep{lecun1998} datasets whether a classification or regression objective performs best for the weakly supervised detection.

Fourth, we evaluate the methods both in MNIST-based detection datasets and in the 3D detection of enlarged perivascular spaces. The MNIST datasets is used as a faster and more controlled experimental setting to study methodological differences between attention map computation methods, optimization objectives, and architectures. We evaluate the best methods in a real-world practical task with clinical relevance: the detection of PVS. The current work is the largest study to date to evaluate automated PVS detection in a large dataset (four regions and 2202 scans) using center locations of PVS.

\begin{figure*}[!t]
\centering
\includegraphics[height=6cm]{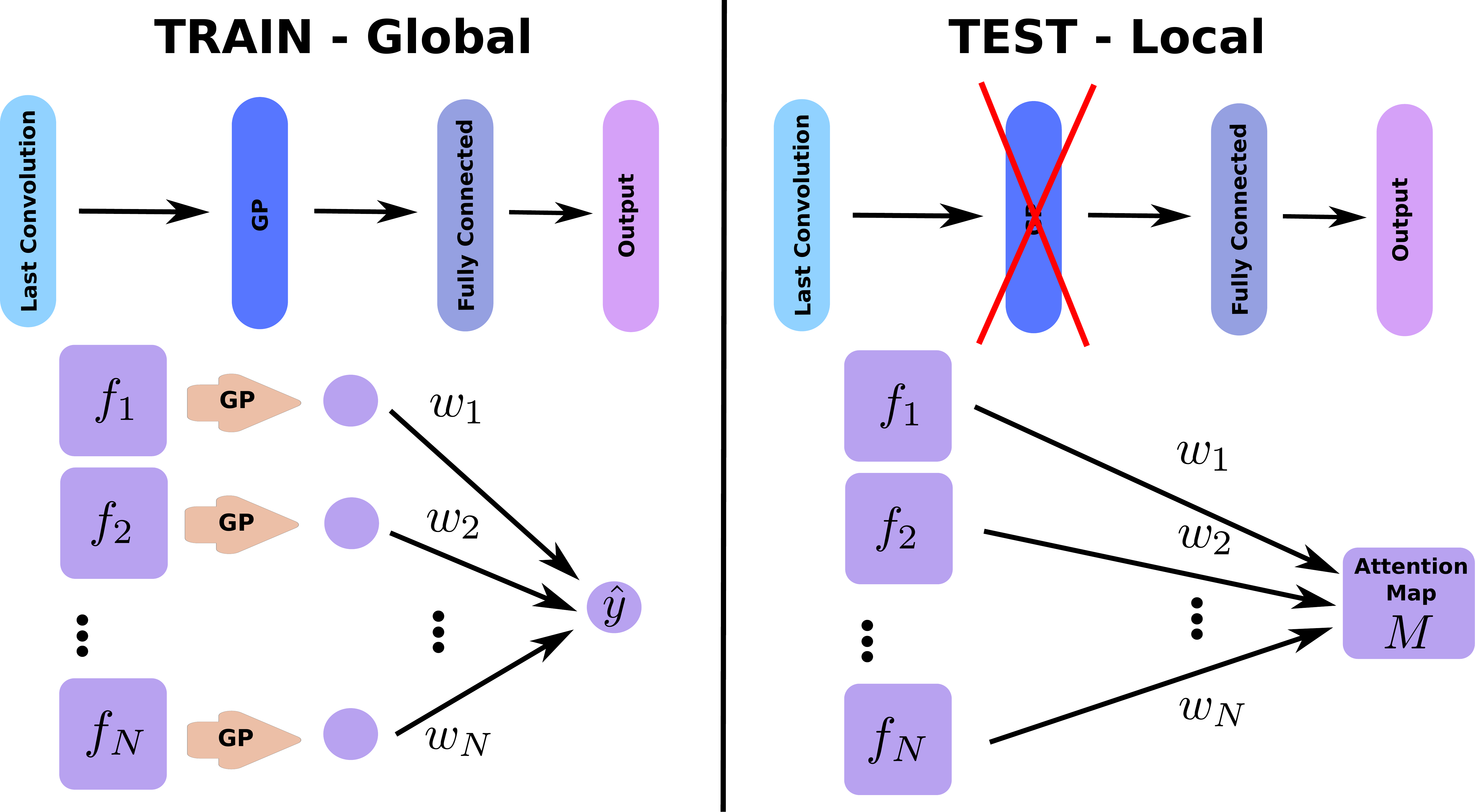}
\caption{\textbf{Principle of CAM methods for regression.} GP stands for Global Pooling. $f_k$ correspond to the feature maps of the last convolutional layer. Disks correspond to scalar values. $w_k$ are the weights of the fully connected layer. Left: the architecture of the network during training. Right: the architecture at inference time, where the global pooling is removed. During training, the network outputs a scalar value which is compared to the image level label to compute the loss and update the network's parameters. During testing, the global pooling layer is removed. Consequently, the network outputs an image. This image is computed as the linear combination of feature maps of the layer preceding the global pooling layer using the weights of the following fully connected layer.}
\label{fig:CAMmethods}
\end{figure*}

\section{Methods}
\label{sec:method}

We implemented seven methods for weakly supervised detection with CNNs: (a) \textit{GP-Unet} (this article) , (b) \textit{GP-Unet no residual} \citep{dubost2017} the first proposed version of GP-Unet, (c) \textit{Gated Attention} \citep{schlemper2018}, (d) \textit{Grad-CAM} \citep{selvaraju2017}, (e) \textit{Grad} \citep{simonyan2014},  (f) \textit{Guided-backpropgation} \citep{springenberg2014}, and (g) an intensity thresholding method for brain datasets only. For all methods, the CNNs are designed to output a single scalar $\hat{y} \in \mathbb{R}$ and are trained with mean squared error using only global labels: the number of occurrences of target objects $y \in \mathbb{N}$. Then for a given input image $I$ the attention map $M$ is computed at inference time. Below, we detail the computation of these attention maps for each method.

\subsection{Computation of the attention maps}

\subsubsection{CAM methods}
The principle of all CAM methods is to use the feature maps -- or activation maps -- of the network to compute attention maps. CAM methods usually exploit the feature maps of the last convolutional layer of the network, as they are expected to be more closely related to the target prediction than feature maps of intermediate layers. \cite{zhou2016} first proposed to introduce a global pooling layer after the last convolution. The global pooling layer projects each feature map $f_{k}$ to a single neuron, resulting in a vector of $N$ scalar values, where $N$ is the number of feature maps $f_{k}$ in the last layer. The global pooling layer is followed by a fully connected layer to a number of neurons corresponding to the number of classes (for classification), or to a single neuron representing the output $\hat{y} \in \mathbb{R}$ (for regression). 
The network can then be trained with image-level labels using, for example, a cross-entropy or mean squared error loss function. During inference the global pooling layer can be removed, and the attention map is then computed as a linear combination of the feature maps $f_{k}$ (before global pooling) using the weights of the fully connected layer $w_{k}$:

\begin{equation}
\label{eq:CAM}
M_{CAM} = \sum_{k}^{N}w_{k}f_{k}.
\end{equation}

The computation of CAM attention maps is illustrated in Figure \ref{fig:CAMmethods}.

\paragraph{GP-Unet}
In the approach by \cite{zhou2016} the attention map is computed from the last feature maps of the network, which are often downsampled with respect to the input image due to pooling layers in the network. To alleviate this problem, we use the same principle with the architecture of a segmentation network (U-net from \cite{ronneberger2015}), i.e. with an upsampling path, where the feature maps $f_{k}$ of the last convolution layer - before global pooling (GP) - have the same size as the input image $I$ (see architectures in Figure \ref{fig:archs} and section \ref{sec:arch}). The attention maps are still computed with Equation \ref{eq:CAM}.

\paragraph{GP-Unet no residual}

In our earlier work, we proposed another version of GP-Unet \citep{dubost2017} based on a deeper architecture without residual connections (see architectures in Figure \ref{fig:archs} and section \ref{sec:arch}). Experiments showed that such deep architecture was not needed \citep{dubost2019}, and could slow the optimization. We refer to this approach as \textit{GP-Unet no residual} in the rest of the paper.
To detect hyperintense brain lesions in MRI data \cite{dubost2017} also rescaled the attention map values to $[0,1]$ and summed them pixel-wise with rescaled image intensities. This is not needed in the new version of GP-Unet above because residual connections between the input and output of two successive convolutional layers allow the network to learn this operation. 

\paragraph{Gated Attention}
While we proposed to upsample and concatenate features maps of different scales \citep{dubost2017} as advised for segmentation networks by \cite{ronneberger2015}, \cite{schlemper2018} proposed instead a more complex gated attention mechanism to combine information from different scales. This gated attention mechanism relies on attention units -- also called attention gates -- that compute soft attention maps and use these maps to mask irrelevant information in the feature maps. Here, global pooling is applied at every scale $s$ and the results are directly linked to the output by a fully connected layer aggregating information across scales. \cite{schlemper2018} proposed three aggregation strategies: concatenation, deep supervision \citep{lee2015}, and fine-tuning by training the network for each scale separately. With the fine tuning strategy, the authors reached a slightly higher performance than concatenation and deep supervision. For the sake of simplicity, we employed the concatenation strategy in our experiments. See Figure \ref{fig:archs} for an illustration of the architectures of Gated Attention and of GP-Unet. The attention maps $M_{Gated}$ of the gated attention mechanism method are computed as:
\begin{equation}
\label{eq:gated}
M_{Gated} = \sum_{s}\sum_{k}^{N_{s}}w_{k}^{s}f_{k}^{s},
\end{equation}
where $w_{k}^{s}$ are the weights of the last fully connected layer for the neurons computed from the feature maps $f_{k}^{s}$ at scale $s$.

\paragraph{Grad-CAM}
Finally, Grad-CAM \citep{selvaraju2017} is a generalization of CAM \cite{zhou2016} to any network architecture. The computation of the attention map is similar to Equation \ref{eq:CAM}, but instead of the weights $w_{k}$, uses new weights $\alpha_{k}$ in the linear combination. The weights $\alpha_{k}$ are computed with the backpropagation algorithm. With this technique the global pooling layer is not needed anymore, and attention maps can be computed from any layer in any network architecture. More precisely, each weight $\alpha_{k}$ is computed as the average over all voxels of the derivative of the output $\hat{y}$ with respect to the feature maps $f_{k}$ of the target convolution layer. In our case, we use the feature maps of the last convolution layer preceding global pooling, and the weights are computed as:
\begin{equation}
\alpha_{k} = \frac{1}{Z} \sum \frac{\partial \hat{y}}{\partial f_{k}},
\end{equation}
where $Z$ is the number of voxels in the feature map $f_{k}$.
The attention map $M_{Grad-CAM}$ is then computed as a linear combination of the feature maps weighted by the $\alpha_{k}$, and upsampled with linear interpolation to compensate the maxpooling layers:
\begin{equation}
\label{eq:grad-CAM}
M_{Grad-CAM} = \sum_{k}^{N}\alpha_{k}f_{k}.
\end{equation}
In their original work, \cite{selvaraju2017} proposed to compute attention maps from any layer in the network. While this approach has the advantage of generating several explanations for the network's behavior, choosing which layer should be used to compute the global attention of network becomes less obvious and objective. In our experiments, we observed that attention maps computed from the first layers of the network highlight large brain structures, and are not helpful for the detection tasks. To be more comparable to the other approaches, we used the feature maps $f_{k}$ of the last convolution layer.

\subsubsection{Gradient methods}
\paragraph{Grad}
\cite{simonyan2014} proposed to compute attention maps by estimating the gradient of the output with respect to the input image. Gradients are computed with the backpropagation algorithm. This method highlights pixels for which a small change would affect the prediction $\hat{y}$ by a large amount. The attention map $M_{Grad}$ is computed as
 
\begin{equation}
M_{Grad} = \frac{\partial \hat{y}}{\partial I}.
\end{equation}

\paragraph{Guided-backpropagation}
The attention maps obtained by Grad can highlight fine detail in the input image, but often display noise patterns. This noise mostly results from negative gradients flowing back in the rectified linear unit (ReLU) activations. In theory these negative gradients should relate to negative contributions to the network prediction, in practice they deteriorate attention maps and are believed to interact with positive gradients according to an interference phenomenon \citep{korbar2017}. With the standard backpropagation algorithm, during the backward pass, ReLU nullifies gradients corresponding to negative entries of the bottom data (input of the ReLU coming from the input to the CNN), but not those that have a negative value in the top layer (which precedes the ReLU during the backward pass). \cite{springenberg2014} proposed to additionally mask out the values corresponding negative entries of the top gradient in the ReLU activations. This is motivated by the deconvolution approach, which can been seen as a backward pass through the CNN where the information passes in reverse direction through the ReLU activations \citep{simonyan2014,springenberg2014}. Masking out these negative entries from the top layer effectively clears the noise in the attention maps.

\subsubsection{Intensity method -- for brain datasets only}

PVS appear as hyperintense areas in the T2-weighted images. In some regions -- especially midbrain, and to some extent basal ganglia -- the image intensity can often be discriminative enough and can be used as a crude attention map. We therefore include the raw image intensity as one of the attention maps in our comparison, and, after non-maximum suppression, use the lesion count $n$ predicted using the base architecture (see Section \ref{sec:arch}) to select the threshold.

\subsection{Architectures}
\label{sec:arch}
\begin{figure*}[!t]
\centering
\includegraphics[height=9cm]{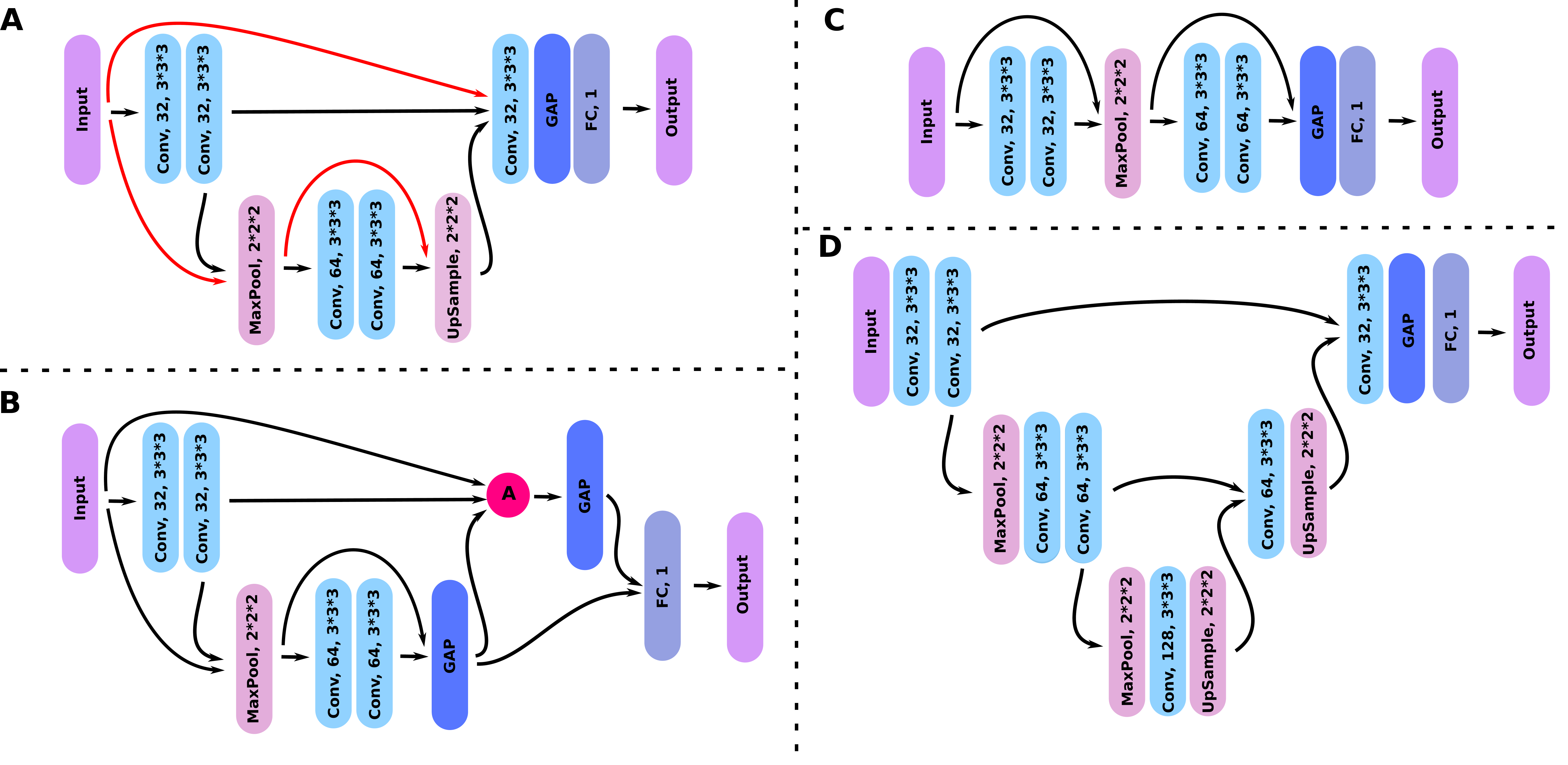}
\caption{\textbf{Architectures.} A is GP-Unet's architecture. B is Gated Attention architecture. C is the base architecture used for Grad, Guided-backpropagation, and Grad-CAM. D is GP-Unet no residual architecture. GAP stand for global average pooling layer, FC for fully connected layer, and A for attention gate. All architectures are detailed in Section \ref{sec:arch}. In architecture A, we showed in red the blockwise skip connections.}
\label{fig:archs}
\end{figure*}
In total, four architectures were implemented to evaluate all six methods. These architectures are illustrated in Figure \ref{fig:archs}. Grad, Guided-backpropagation, and Grad-CAM use the same neural networks (same architecture and weights), but differ in the computation of the attention maps during inference. The other methods require different architectures, and are trained separately. In the following section, we detail the components of each architecture in 3D.

We perform experiments on 2D CNNs for the MNIST dataset and on 3D CNNs for the brain dataset.
The 3D CNNs use 3D convolutional layers with 3x3x3 filters with zero-padding, and 3D maxpooling layers of size 2x2x2. Similarly, the 2D CNNs use 2D convolutional layers with 3x3 filters with zero-padding, and 2D maxpooling layers of size 2x2. The 2D CNNs always use four times fewer features maps than their 3D counterpart to allow faster experimentation. After the last convolution layer, each feature map is projected to a single neuron using global average pooling. These neurons are connected with a fully connected layer to a single neuron indicating the output of network $\hat{y} \in \mathbb{R}$. Rectified linear unit (ReLU) activations are used after each convolution. We use skip connections by concatenating the feature maps of different layers (and not by summing them).

\paragraph{GP-Unet architecture (A in Figure \ref{fig:archs})}

GP-Unet architecture is that of small segmentation network, with an encoder and a decoder part. The architecture starts with two convolutional layers with 32 filters each. The output of these two layers is concatenated with the input. Then follows a maxpooling layer and two convolutional layers with 64 filters each. The feature maps preceding and following these two layers are concatenated. In order to combine of features at different scales, these low dimension feature maps are upsampled, concatenated with features maps preceding the maxpooling layer, and given to a convolutional layers of 32 filters. Then follows a global average pooling layer, from which a fully connected layer maps to the output. 
This architecture is simple (308 705 parameters for the 3D version), fast to train (less than one day on 1070 Nvidia GPU), and allows computing attention maps at the full resolution of the input image.

\paragraph{GP-Unet no residual architecture (D in Figure \ref{fig:archs})}

The architecture of GP-Unet no residual was proposed by \citep{dubost2017}. In this work, we only changed the global pooling layer from maximum to average to make comparisons between methods more meaningful. This network is a segmentation network with a downsampling and upsampling path. The downsampling path has two convolutional layers of 32 filters, a maxpooling layer, two convolutional layers of 64 filters, a maxpooling layer, and one convolutional layer of 128 filters. The upsampling path starts with an upsampling layer, concatenates the upsampled feature maps with the features maps preceding the maxpooling layer in the downsampling path, computes a convolutional layer with 64 filters, and repeat this complete process for the last scale of feature maps, with a convolutional layer of 32 filters. After that, comes the global pooling layer, and fully connected layer to a single neuron.

The difference with architecture (A) \citep{dubost2017} is that the feature maps are downsampled twice instead of once, and that there are no skip connections between sets of two consecutive convolutions (blockwise skip connection in red in Figure \ref{fig:archs}). Consequently, the last convolution layer does not have access to the input image intensities. We believe these residual connections make the design of GP-Unet more flexible than this architecture, by facilitating for instance the network to directly use the input intensities and locally adjust its predictions. This can be crucial for the correct detection of brain lesions.
This architecture has twice more parameters (637 185 parameters for the 3D version) than that of GP-Unet.

\paragraph{Gated Attention architecture (B in Figure \ref{fig:archs})}
We adapted the architecture of the Gated Attention network proposed by \cite{schlemper2018} to make it more comparable to the other approaches presented in the current work. Here, the Gated Attention architecture is the same as GP-Unet architecture (A) except for two differences: to merge the feature maps between the two different scales, instead of upsampling, concatenation and convolution, we use the attention gate as described by \cite{schlemper2018}. The other difference is that, in this architecture (B), the downsampled feature maps are also projected to single neurons with global pooling. The neurons corresponding to the two different scales are then aggregated (using concatenation) and connected to the single output neuron with a single fully connected layer.
This architecture has 198 580 parameters for the 3D version.

The attention gate computes a normalized internal attention map. In their implementation, \cite{schlemper2018} proposed a custom normalization to prevent the attention map from becoming too sparse. We did not experience such problems and opted for the standard sigmoid normalization.

Similarly to GP-Unet, Gated Attention computes attention maps at the resolution of the input image. However it combines multi-level information with a more complex process than GP-Unet.

\paragraph{Base architecture (C in Figure \ref{fig:archs})}

The network architecture used for Grad, Guided-backpropagation, and Grad-CAM is kept as similar as possible to that of GP-Unet for better comparison of methods. It starts with two convolutional layers with 32 filters each. The output of these two layers is concatenated with the input. Then follows a maxpooling layer and two convolutional layers with 64 filters each. The output of these two layers is concatenated with the feature maps following the maxpooling layer, and is given directly to the global average pooling layer. In other words, we apply global pooling to the original image (after maxpooling) and the feature maps after the second convolution at each scale - so on 1+32+64 feature maps. This architecture has shown competitive performance on different types of problems in our experiments (eg. in brain lesions in \citep{dubost2018}). With this architecture, unlike GP-Unet, Grad-CAM produces attention maps at a resolution twice smaller than that of the input image, and could miss small target objects.
This architecture has 196 418 parameters for the 3D version.

\section{Experiments}

In this work, we compare our proposed method to five weakly supervised detection methods. 
We use the MNIST datasets \citep{lecun1998} to compare regression against classification for weak supervision. We compared performance of the different methods -- using regression objectives -- on weakly supervised lesion detection in a large brain MRI dataset.

\subsection{MNIST Datasets}
\label{sec:dataMNIST}
We construct images as a grid of 7 by 5 randomly sampled MNIST digit images.  Examples are shown in Figures \ref{fig:MNISTallMethods} and \ref{fig:MNIST}. Each digit is uniformly drawn from the set of all training/validation/testing digits, hence with a probability 0.1 to be a target digit $d$.
To avoid class imbalance, we adapt the dataset to each target digit $d$ by sampling 50\% of images with no occurrence of $d$, and 50\% of images with at least one occurence of $d$, resulting in ten different datasets.

\subsection{Brain Datasets}
\label{sec:dataBrain}

Brain MRI was performed on a 1.5-Tesla MRI scanner (GE-Healthcare, Milwaukee, WI, USA) with an eight-channel head coil to obtain 3D T2-contrast magnetic resonance scans. The full imaging protocol has been described by \cite{ikram2015}. In total, our dataset contains 2202 brain scans, each scan being acquired from a different subject.

\begin{figure}[]
\centering
\includegraphics[height=6cm]{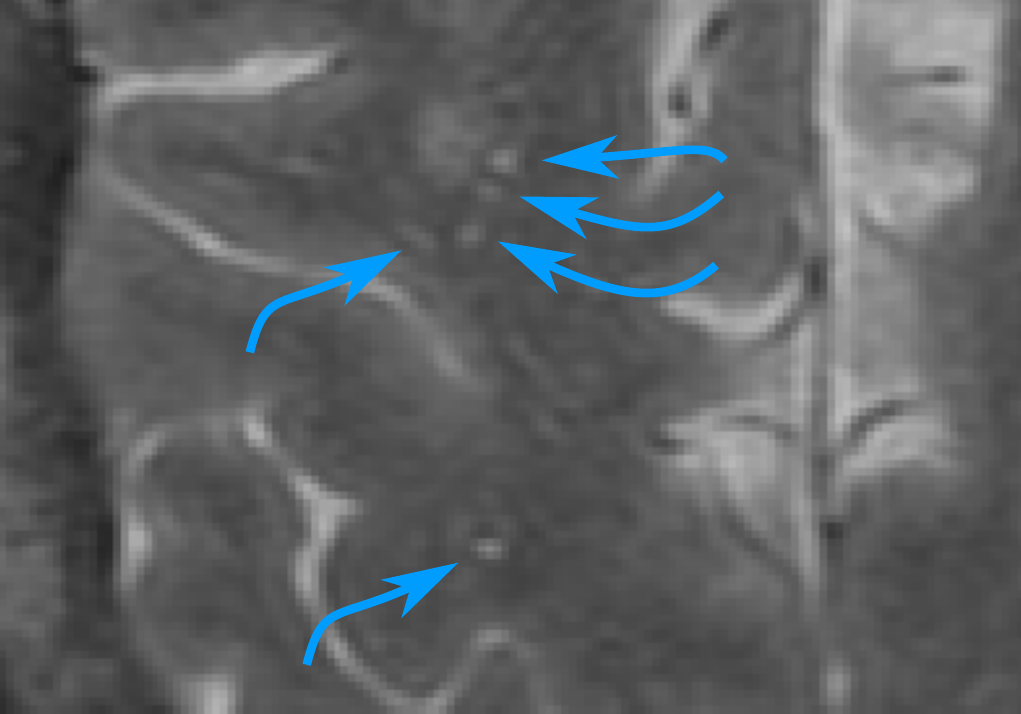}
\caption{\textbf{Examples of PVS in the centrum semiovale.} This is a crop of a T2-weighted image in axial view. PVS are indicated with blue arrows.}
\label{fig:PVSexample}
\end{figure}

An expert rater annotated PVS in four brain regions: in the complete midbrain and hippocampi, and in a single slice in axial view in the basal ganglia (the slice showing the anterior commissure) and the centrum semiovale (the slice 10 cm above the top of the lateral ventricle). The annotation protocol follows the guidelines by \cite{adams2014} and \cite{Adams2013} for visual scoring of PVS, with the difference that \cite{adams2014} only counted the number of PVS, while in the current work, all PVS have been marked with a dot in their center. Figure \ref{fig:PVSexample} shows examples of PVS in the centrum semiovale.

\subsection{Aim of the experiments}

In the MNIST datasets, the objective is to detect all occurrences of a target digit $d$. During optimization, the regression objective is to count the number of occurrences of $d$, while the classification objective is to detect the presence of at least one occurence of $d$.

In the experiments on 3D brain MRI scans, the objective is to detect enlarged perivascular spaces (PVS) in the four brain regions described in section \ref{sec:dataBrain}. For these datasets we investigate only regression neural networks. These networks are optimized using the number of annotated PVS in the region of interest as the weak global label, as proposed in our earlier work \cite{dubost2018}. The location of PVS are only used for the evaluation of the detection during inference.

\subsection{Preprocessing}
\label{sec:preProcess}
\paragraph{MNIST data}
We scale the image intensity values in the MNIST grid images between zero and one to ease the learning process.

\paragraph{Brain scans}
We first apply the FreeSurfer multi-atlas segmentation algorithm \citep{Desikan2006} to locate and mask the midbrain, hippocampi, basal ganglia and centrum semiovale in each scan. For each region, we then extract a fixed volume centered on the center of mass of the region. For midbrain (88x88x11 voxels), hippocampi (168x128x84 voxels) and basal ganglia (168x128x84 voxels) these cropped volumes contain the full region. The centrum semiovale is too large to fit in the memory of our GPU (graphics processing unit), so for this region we only extract the slices surrounding the slice that was scored by the expert rater (250x290x14 voxels). Consequently, we apply a smooth region mask to nullify values corresponding to other brain regions. Finally, we scale the intensity values between zero and one to ease the learning process.
The preprocessing and extraction of brain regions is presented in more details in previous work \citep{dubost2018}.

\subsection{Training of the networks}

All regression networks are optimized with Adadelta \citep{zeiler2012} to minimize the mean squared error between their prediction $\hat{y} \in \mathbb{R}$ and the ground truth count $y \in \mathbb{N}$. The classification networks in our MNIST experiments were optimized with Adadelta and the binary cross-entropy loss function.

Weights of the convolution filters and fully connected layers are initialized from a Gaussian distribution with zero mean and unit variance, and biases are initialized to zero.

A validation set is used to prevent over-fitting. The optimization is stopped at least 100 epochs after the validation loss stopped decreasing. We select the model with the lowest validation loss. For the MNIST datasets, the models are trained on a set of 500 images (400 for training and 100 for validation). For the brain datasets, the models are trained on a set of 1202 scans (1000 for training and 202 for validation).
During training, we use on-the-fly data augmentation with a random combination of random translations of up to 2 pixels in all directions, random rotations up to 0.2 radians in all directions, and random flipping in all directions.
For the MNIST datasets, the batch size was set to 64. For the brain datasets, because of GPU memory constraints, the networks are trained per sample: each mini-batch contains a single 3D image. As the convergence can be slow in some datasets, we first trained the networks on the smallest and easiest region (midbrain), and fine-tune the parameters for the other regions, similarly to \cite{dubost2018}.

We implemented our algorithms in Python in Keras \citep{chollet2015} with TensorFlow as backend, and ran the experiments on a Nvidia GeForce GTX 1070 GPU and Nvidia Tesla K40 \footnote{We used computing resources provided by SurfSara at the Dutch Cartesius cluster.}. The average training time was one day.

\begin{figure*}[!t]
\centering
\includegraphics[width=\textwidth]{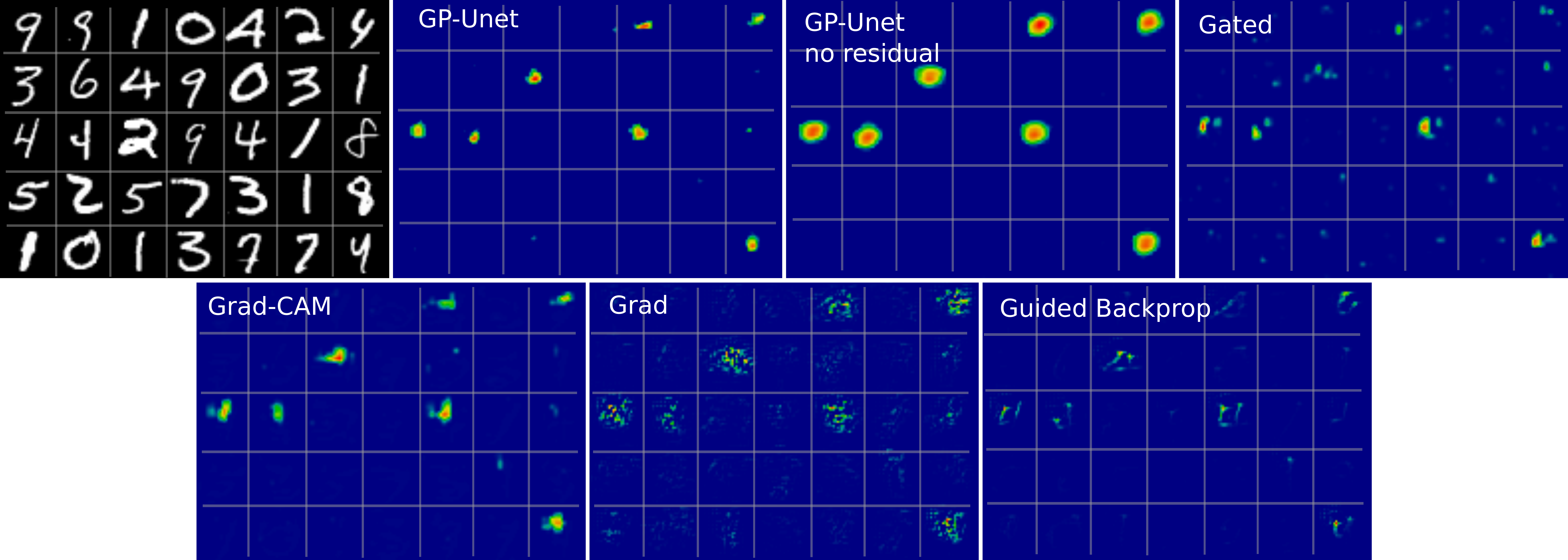}
\caption{\textbf{Examples of attention maps of the different weakly supervised detections methods for the detection of digit 4.} Top-left: MNIST image. All methods with optimized with regression objectives.}
\label{fig:MNISTallMethods}
\end{figure*}

\begin{figure*}[!t]
\centering
\includegraphics[width=\textwidth]{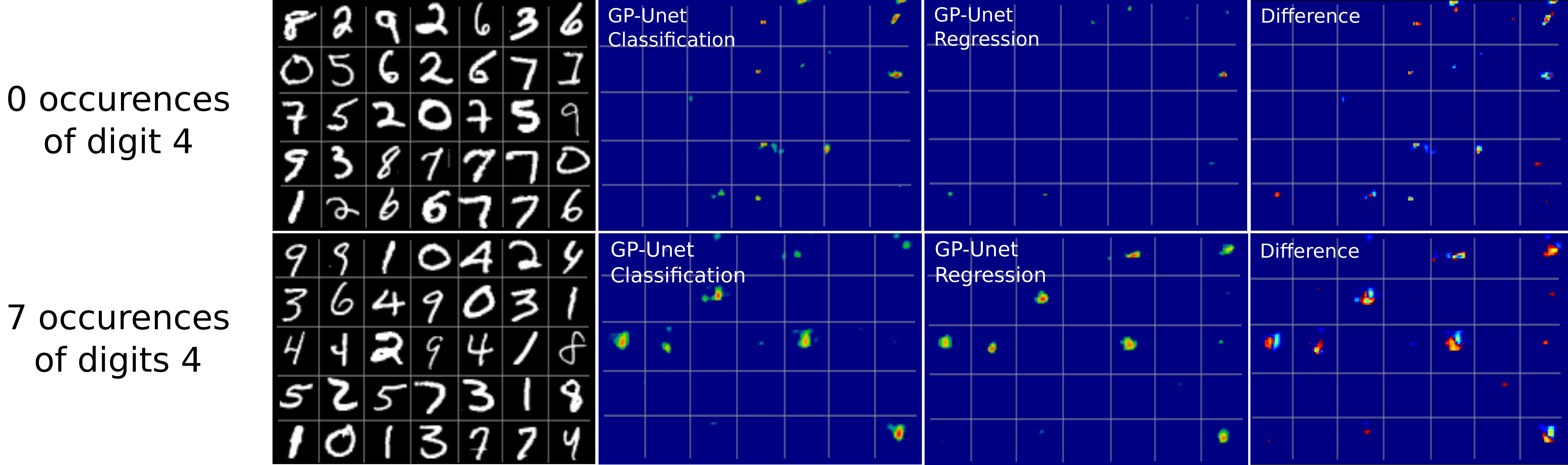}
\caption{\textbf{Examples of attention maps of GP-Unet for the detection of digit 4 and optimized with classification and regression objectives.} Left: MNIST image, middle: attention map generated from a classification network, right: attention map generated from a regression network. The first row displays an image without digit 4. The second row displays an image with seven occurences of the digit 4. For the classification method, in the first row we notice more false positives than for the regression method. On the second row, the two digits 4 at the top are less highlighted than the other digits 4 in the image. It is not the case for the regression attention map. This observation supports the hypothesis that attention maps computed from classification objectives tend to focus more on the most obvious occurence of the target object, instead of equally focusing on all occurrences. On the right, we show the difference between the attention maps for regression and classification.}
\label{fig:MNIST}
\end{figure*}

\subsection{Negative values in attention maps}
\label{sec:neg}

Attention maps can have negative values, which meaning can differ for CAM methods and gradient methods. For CAM methods, negative values could highlight objects in the image which presence is negatively associated with the target objects.
For gradient methods, they correspond to areas where increasing the intensity would decrease the predicted count (or where decreasing the intensity would increase the predicted count, these are the same areas). 

For image understanding, keeping negative values in attention maps seems most appropriate as the purpose is to discover which parts of the image contributed either negatively or positively to the prediction, and how a change in their intensity could affect the prediction. For detection, the purpose is to find to find all occurrences of the target object in the image and ignore other objects. In the literature, two approaches have been proposed to handle negative values for object detection: either setting them to zero, or taking the absolute value. CAM methods \citep{zhou2016,selvaraju2017} nullify negative values of the attention maps to mimic the behavior of ReLU activations. Gradient methods \citep{simonyan2014, springenberg2014} focus on the magnitude of the derivative and thus compute the absolute value.

In our case, we aim to solve a detection problem in datasets where the target objects are among the highest intensity values in the image. For gradient methods, this implies that negatives values in the attention maps do not indicate the location of the target object in our case. We can therefore ignore negative values, and decided to nullify them. For CAM methods, we follow the recommendation of the literature, and also nullify negative values in attention maps. Consequently, we nullified negative values for all methods. Nullifying negative values actually only impacts the visualization of the attention maps, and not the detection metrics, as we select only candidates with highest values in the attention maps (Section \ref{sec:postproc}). On the contrary taking the absolute value could increase the number of detections and would impact our detection metrics.

\subsection{Performance evaluation}
\label{sec:postproc}

The output of all weakly-supervised detection methods presented in Section \ref{sec:method} are attention maps. We still need to obtain the coordinates of the detections, and evaluate the matching with the ground truth.

After setting negative values to zero (Section \ref{sec:neg}), we apply non-maximum suppression on the attention maps using a 2D (MNIST, centrum semiovale and basal ganglia) or 3D (hippocampi and midbrain) maximum filter of size 6 voxels (which corresponds to 3 mm in axial plane, the maximum size for PVS as defined by \cite{Adams2013} -- we used the same value for the MNIST datasets) with 8 neighborhood in 2D or 26 neighborhood in 3D. This results in a set of candidates that we order according to their value in the attention map. The candidates with highest values are considered the most likely to be the target object. 

For the basal ganglia and the centrum semiovale, our dataset does not contain full 3D annotations, but only provides annotations for a single 2D slice per scan (see Section \ref{sec:dataBrain}). 
As annotations were only available in a single slice, we evaluated the attention maps only in the annotated slice, although we can compute attention maps for the complete volume of these regions.
For our evaluation we extract the corresponding 2D slice from the attention map prior to post-processing and compute the metrics only for this slice. In case no lesion was annotated, we selected the middle slice of the attention map as a reasonable approximation of the rated slice.

As we aim to solve a detection problem, we need to quantify the matching between two sets of dots: the annotators dots, and the algorithms’ predictions. We used the Hungarian algorithm \citep{kuhn1955} to create an optimal one-to-one match between each detected lesion or digit to the closest annotation in the ground truth. For the brain dataset, we counted a positive detection if a detection was within at most 6 voxels from the corresponding point in the ground truth. This corresponds to the maximum diameter of PVS in the axial view, as defined in \cite{Adams2013}. For the MNIST datasets, we counted a positive detection if a detection fell inside the 28*28 pixels wide original MNIST image of the target digit.

As the algorithms output candidates with confidence scores, we can compute free-response receiver operating characteristic (FROC) curves \citep{bandos2009} that show the trade-off between high sensitivity and the number of false positives, in our case more precisely the average number of false positives per scan (FPavg). To draw these curves, we varied the number of selected candidates.  For each network in our experiments, we report the area under the FROC curve (FAUC) computed from 0 to 5 FPavg for MNIST and from 0 to 15 FPavg for brain lesion detection. We also show the standard deviation of the FAUC, computed by bootstrapping the test set.

In addition to the attention maps, the regression networks also predict the number of target objects in the image. For the detection of brain lesions, we use this predicted count rounded to an integer $n$ to select the top-$n$ candidates with highest scores, and compute the corresponding sensitivity and FPavg,
and the average number of false negative per scan (FNavg).
For statistical significance of difference of FAUCs, we performed a bootstrap hypothesis testing and consider statistical significance for p-value lower than 0.05. For FPavg, FNavg and Sensitivity we performed Wilcoxon tests using p-value lower than 0.05.

\subsection{Intra-rater variability of the lesion annotations}
\label{sec:intrarater}
Intra-rater variability has been measured in each region using a separate set of 40 MRI scans acquired and annotated with the same protocol. The rater annotated PVS twice in each scan with two weeks of interval, and in a different random order.

\begin{figure*}[!t]
\centering
\includegraphics[height=12cm]{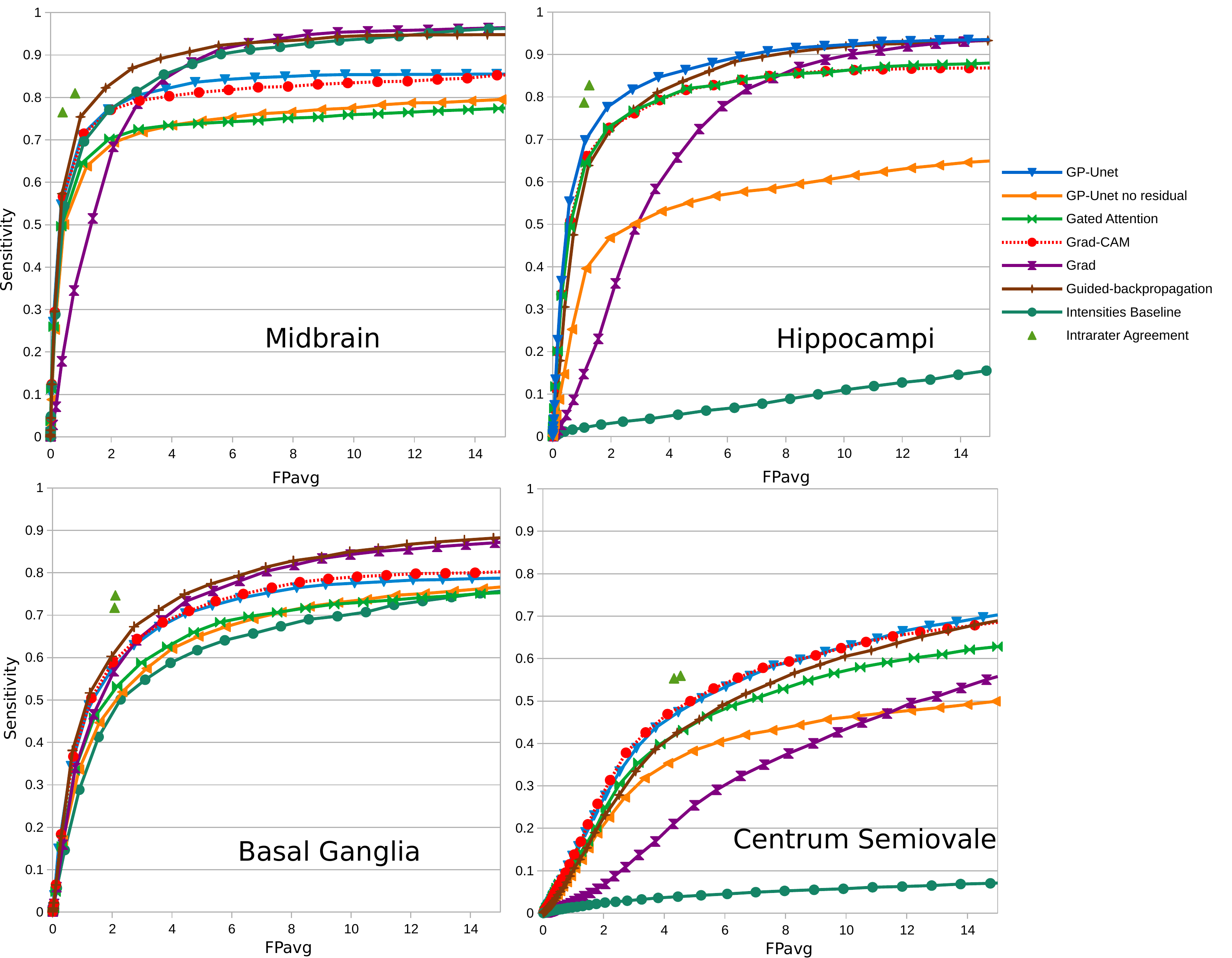}
\caption{\textbf{FROC curves of enlarged perivascular spaces detection in the brain MRI in four different regions.} The average number of false positives per scan is displayed on the x-axis, and the sensitivity on the y-axis. Axes have been rescaled for better visibility. The green triangles indicate intra-rater agreement (on a smaller set) as described in Section \ref{sec:intrarater} .}
\label{fig:froc}
\end{figure*}

To compute the sensitivity and FPavg for the Intra-rater variability, one of the two series of annotations has to be set as reference to define true positives, positives and false positives. We successively set the first and second series of annotations as reference, leading to two different results.
All results for all regions are displayed next to the FROC curves in Figure \ref{fig:froc}.

\section{Results}
\label{sec:results}

\begin{figure*}[!t]
\centering
\includegraphics[height=5.5cm]{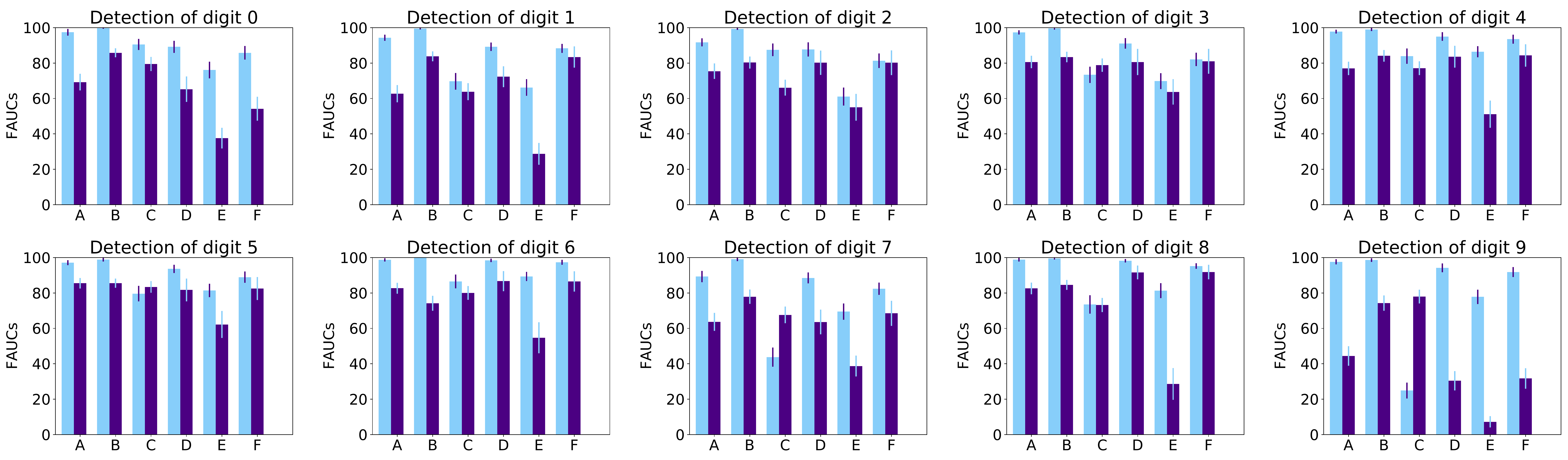}
\caption{\textbf{FAUCs (Section \ref{sec:postproc}) on the MNIST dataset for all methods.} Each subplot corresponds to the detection of a different digit. Results for regression networks are displayed in light blue, and results for classification networks are displayed in indigo. FAUCs are displayed with standard deviations computed by bootstrapping the test set. A is GP-Unet, B GP-Unet no residual, C Gated Attention, D Grad-CAM, E Grad and F Guided-backpropagation.}
\label{fig:faucMNIST}
\end{figure*}

\begin{table*}[h]
\caption{\textbf{FAUCs for the detection of brain lesions.} To compute the these FAUCs, we integrate the FROC (Figure \ref{fig:froc}) between 0 and 15 (Section \ref{sec:postproc}). The best performance in each region is indicated in bold.}
\resizebox{\textwidth}{!}{%
\begin{tabular}{cccccccc}
\hline
                                                             & \begin{tabular}[c]{@{}c@{}}GP-Unet\\ (this paper)\end{tabular} & \begin{tabular}[c]{@{}c@{}}GP-Unet no residual\\ \cite{dubost2017}\end{tabular} & \begin{tabular}[c]{@{}c@{}}Gated Attention\\ \cite{schlemper2018}\end{tabular} & \begin{tabular}[c]{@{}c@{}}Grad-CAM\\ \cite{selvaraju2017}\end{tabular} & \begin{tabular}[c]{@{}c@{}}Grad\\ \cite{simonyan2014}\end{tabular} & \begin{tabular}[c]{@{}c@{}}Guided-backprop\\ \cite{springenberg2014}\end{tabular} & \begin{tabular}[c]{@{}c@{}}Intensities\\ Section \ref{sec:baseline}\end{tabular}\\ \hline
Midbrain & 81.5 (80.1 - 82.8) & 73.4 (72.0 - 74.8) & 72.7 (71.1 - 74.4) & 79.8 (78.5 - 81.1) & 84.5 (83.5 - 85.4) & \textbf{89.2 (88.3 - 90.2)} & 87.1 (86.1 - 88.1) \\
Hippocampi & \textbf{85.8 (84.8 - 86.7)} & 55.1 (53.5 - 56.7) & 80.2 (79.1 - 81.3) & 80.1 (78.9 - 81.3) & 71.5 (70.4 - 72.6) & 83.3 (82.2 - 84.3) & 8.3 (7.5 - 9.0)    \\
Basal Ganglia  & 69.6 (68.1 - 71.2) & 64.4 (63.0 - 65.9) & 64.8 (63.4 - 66.4) & 70.6 (69.3 - 72.0) & \textbf{73.5 (72.2 - 74.9)} & \textbf{75.6 (74.3 - 76.8)} & 61.7 (59.9 - 63.5) \\
\begin{tabular}[c]{@{}c@{}}Centrum \\ Semiovale\end{tabular} & \textbf{51.3 (50.1 - 52.6)} & 37.9 (36.8 - 39.2) & 46.2 (45.0 - 47.5) & \textbf{51.5 (50.2 - 52.7)} & 31.9 (30.7 - 33.2) & 48.1 (46.9 - 49.3) & 4.7 (4.2 - 5.2)    \\
\hdashline[1.5pt/5pt]
Average & 72.0 +/- 13.3       & 57.7 +/- 13.1       & 66.0 +/- 12.7       & 70.5 +/- 11.6       & 65.4 +/- 19.9       & 74.1 +/- 15.7       & 40.5 +/- 35.2       \\ \hline
\label{table:fauc}
\end{tabular}
}
\end{table*}

\begin{table*}[h]
\caption{\textbf{Sensitivity in the brain datasets.} Best performance are indicated in bold.}
\resizebox{\textwidth}{!}{%
\begin{tabular}{cccccccc}
\hline
                                                             & \begin{tabular}[c]{@{}c@{}}GP-Unet\\ (this paper)\end{tabular} & \begin{tabular}[c]{@{}c@{}}GP-Unet no residual\\ \cite{dubost2017}\end{tabular} & \begin{tabular}[c]{@{}c@{}}Gated Attention\\ \cite{schlemper2018}\end{tabular} & \begin{tabular}[c]{@{}c@{}}Grad-CAM\\ \cite{selvaraju2017}\end{tabular} & \begin{tabular}[c]{@{}c@{}}Grad\\ \cite{simonyan2014}\end{tabular} & \begin{tabular}[c]{@{}c@{}}Guided-backprop\\ \cite{springenberg2014}\end{tabular} & \begin{tabular}[c]{@{}c@{}}Intensities\\ Section \ref{sec:baseline}\end{tabular}\\ \hline
Midbrain & 71.1 (69.5 - 72.7) & 63.8 (62.1 - 65.5) & 64.6 (62.8 - 66.3) & 71.5 (69.8 - 73.1) & 51.5 (49.6 - 53.3) & \textbf{75.4 (73.8 - 77.0)} & 69.6 (67.9 - 71.4) \\
Hippocampi & \textbf{69.8 (68.2 - 71.3)} & 46.8 (45.2 - 48.4) & 64.6 (62.9 - 66.2) & 66.1 (64.5 - 67.6) & 36.1 (34.5 - 37.6) & 63.8 (62.2 - 65.5) & 4.2 (3.6 - 4.8)    \\
Basal Ganglia  & 56.8 (55.0 - 58.5) & 51.9 (50.1 - 53.6) & 53.3 (51.6 - 55.0) & 58.9 (57.2 - 60.6) & 56.8 (55.1 - 58.5) & \textbf{60.3 (58.6 - 62.0)} & 50.1 (48.3 - 52.0) \\
\begin{tabular}[c]{@{}c@{}}Centrum \\ Semiovale\end{tabular} & 50.6 (49.3 - 52.0) & 42.0 (40.7 - 43.4) & 48.8 (47.5 - 50.2) & \textbf{53.0 (51.6 - 54.3)} & 35.0 (33.9 - 36.1) & 49.0 (47.7 - 50.3) & 5.7 (5.2 - 6.3)    \\
\hdashline[1.5pt/5pt]
Average & 62.1 +/- 8.7        & 51.1 +/- 8.1        & 57.8 +/- 6.9        & 62.4 +/- 7.0        & 44.8 +/- 9.5        & 62.1 +/- 9.4        & 32.4 +/- 28.3       \\ \hline
\label{table:sensitivity}
\end{tabular}
}
\end{table*}

\begin{table*}[h]
\caption{\textbf{Average number of false positives per scan in the brain datasets.} Best performances are indicated in bold.}
\resizebox{\textwidth}{!}{%
\begin{tabular}{cccccccc}
\hline
                                                             & \begin{tabular}[c]{@{}c@{}}GP-Unet\\ (this paper)\end{tabular} & \begin{tabular}[c]{@{}c@{}}GP-Unet no residual\\ \cite{dubost2017}\end{tabular} & \begin{tabular}[c]{@{}c@{}}Gated Attention\\ \cite{schlemper2018}\end{tabular} & \begin{tabular}[c]{@{}c@{}}Grad-CAM\\ \cite{selvaraju2017}\end{tabular} & \begin{tabular}[c]{@{}c@{}}Grad\\ \cite{simonyan2014}\end{tabular} & \begin{tabular}[c]{@{}c@{}}Guided-backprop\\ \cite{springenberg2014}\end{tabular} & \begin{tabular}[c]{@{}c@{}}Intensities\\ Section \ref{sec:baseline}\end{tabular}\\ \hline
Midbrain & \textbf{1.03 (0.99 - 1.07)} & 1.19 (1.15 - 1.24) & \textbf{1.04 (0.99 - 1.09)} & 1.10 (1.05 - 1.15) & 1.40 (1.34 - 1.45) & \textbf{0.99 (0.94 - 1.03)} & 1.11 (1.06 - 1.15)  \\
Hippocampi & \textbf{1.12 (1.06 - 1.17)} & 1.96 (1.88 - 2.03) & \textbf{1.13 (1.06 - 1.19)} & \textbf{1.16 (1.10 - 1.22)} & 2.16 (2.06 - 2.25) & 1.23 (1.16 - 1.29) & 3.34 (3.22 - 3.45)  \\
Basal Ganglia  & \textbf{1.95 (1.88 - 2.01)} & 2.33 (2.27 - 2.39) & 2.16 (2.10 - 2.23) & 2.02 (1.95 - 2.09) & 2.06 (1.98 - 2.13) & \textbf{1.98 (1.91 - 2.04)} & 2.28 (2.21 - 2.35)  \\
\begin{tabular}[c]{@{}c@{}}Centrum \\ Semiovale\end{tabular} & \textbf{5.24 (5.04 - 5.43)} & 6.66 (6.46 - 6.86) & 6.23 (6.02 - 6.44) & 5.63 (5.44 - 5.82) & 7.30 (7.03 - 7.57) & 5.92 (5.71 - 6.12) & 9.91 (9.62 - 10.21) \\
\hdashline[1.5pt/5pt]
Average & 2.33 +/- 1.71      & 3.04 +/- 2.13      & 2.64 +/- 2.12      & 2.48 +/- 1.86      & 3.23 +/- 2.37      & 2.53 +/- 1.99      & 4.16 +/- 3.41       \\ \hline
\label{table:fpavg}
\end{tabular}
}
\end{table*}

\begin{table*}[h]
\caption{\textbf{Average number of false negatives per scan in the brain datasets.} Best performances are indicated in bold.}
\resizebox{\textwidth}{!}{%
\begin{tabular}{cccccccc}
\hline
                                                             & \begin{tabular}[c]{@{}c@{}}GP-Unet\\ (this paper)\end{tabular} & \begin{tabular}[c]{@{}c@{}}GP-Unet no residual\\ \cite{dubost2017}\end{tabular} & \begin{tabular}[c]{@{}c@{}}Gated Attention\\ \cite{schlemper2018}\end{tabular} & \begin{tabular}[c]{@{}c@{}}Grad-CAM\\ \cite{selvaraju2017}\end{tabular} & \begin{tabular}[c]{@{}c@{}}Grad\\ \cite{simonyan2014}\end{tabular} & \begin{tabular}[c]{@{}c@{}}Guided-backprop\\ \cite{springenberg2014}\end{tabular} & \begin{tabular}[c]{@{}c@{}}Intensities\\ Section \ref{sec:baseline}\end{tabular}\\ \hline
Midbrain & 0.77 (0.71 - 0.83) & 0.98 (0.91 - 1.05) & 0.94 (0.87 - 1.00) & 0.77 (0.71 - 0.82) & 1.06 (1.00 - 1.12) & \textbf{0.65 (0.60 - 0.71)} & 0.77 (0.72 - 0.83)  \\
Hippocampi & \textbf{1.14 (1.07 - 1.22)} & 2.12 (2.01 - 2.23) & 1.33 (1.25 - 1.41) & 1.32 (1.24 - 1.41) & 2.32 (2.21 - 2.43) & 1.39 (1.31 - 1.47) & 3.50 (3.36 - 3.64)  \\
Basal Ganglia  & 2.00 (1.85 - 2.14) & 2.11 (1.97 - 2.25) & 2.08 (1.94 - 2.21) & \textbf{1.92 (1.78 - 2.06)} & 1.96 (1.82 - 2.09) & \textbf{1.88 (1.74 - 2.01)} & 2.18 (2.03 - 2.33)  \\
\begin{tabular}[c]{@{}c@{}}Centrum \\ Semiovale\end{tabular} & 5.83 (5.50 - 6.17) & 6.67 (6.30 - 7.03) & 5.98 (5.64 - 6.32) & \textbf{5.63 (5.30 - 5.96)} & 7.30 (6.92 - 7.68) & 5.92 (5.58 - 6.26) & 9.91 (9.44 - 10.38) \\
\hdashline[1.5pt/5pt]
Average & 2.44 +/- 2.01      & 2.97 +/- 2.18      & 2.58 +/- 2.00      & 2.41 +/- 1.90      & 3.16 +/- 2.43      & 2.46 +/- 2.04      & 4.09 +/- 3.50       \\ \hline
\label{table:fnavg}
\end{tabular}
}
\end{table*}

\subsection{Regression vs classification objectives - MNIST datasets}
The methods were evaluated on left-out test sets of 500 images, balanced as described in section \ref{sec:dataMNIST}.
Figure \ref{fig:faucMNIST} compares the FAUC of regression and classification networks, for all MNIST digits, and for all methods. 
Additional results such as FROC curves, sensitivity, FPavg and FNavg are 
given in \ref{appendix:resultsMNISTreg} and \ref{appendix:resultsMNISTclassification}.
Overall, regression methods reach a higher detection performance than classification methods. For all digits, regression GP-Unet no residual reaches the best performance. The second best method for all digits is regression GP-Unet. Both GP-Unet regression methods are consistently better than any other method for all digits. Regression Grad-CAM comes third, and regression Guided-backpropagation fourth. Grad and Gated Attention come last. The ordering of best classification methods is different than that of the best (regression) methods: Guided-backpropagation comes first, Grad-CAM second and GP-Unet no residual third. 

Figure \ref{fig:MNISTallMethods} shows an example of the attention maps obtained for all weakly supervised methods optimized with regression objectives. As expected, Grad produces noisy attention maps with many high values, for both classification and regression objectives, and Guided-backpropagation corrects these mistakes. Gradient methods seems to highlight multiple discriminating features of the digit 4 (e.g. its top branches), while CAM methods highlight a single larger, less detailed region. This may suggest that gradients methods may be more suited to weakly supervised segmentation, although judging from the figure, none of the methods seems capable of correctly segmenting digits. 

Figure \ref{fig:MNIST} compares attention maps of GP-Unet optimized with regression and classification. We noticed two interesting differences. First, when the target digit is present on the image, the regression attention map highlights each occurrence of the target digits with a similar intensity, while the classification attention map highlights more strongly the most obvious occurrences of the target digit. Second, when the target digit is not present in the image, contrary to the regression attention map, the classification attention map may highlight many false positives, possibly resulting in a significant drop in the detection performance.

\paragraph{Regression Guided-backpropagation vs Grad.} Regression Guided-backpropagation detects of all digits more accurately than regression Grad. The same comparison holds for classification Guided-backpropagation versus classification Grad. However Regression Grad sometimes performs as well (digits 4, 6, 7) or better (digits 0, 9) than Classification Guided-backpropagation, which underlines the added-value of optimizing weakly supervised detection methods with regression objectives instead of classification objectives.

\subsection{Variations of the architecture of GP-Unet - MNIST datasets}
In this section we studied the influence of the skip connections between sets of two consecutive convolutions (blockwise skip connections, in red in Figure \ref{fig:archs}) in GP-Unet’s architecture and the influence of the type of global pooling in GP-Unet’s architecture on the detection performance. 
Removing the blockwise skip connections did not make the detection worse for most digits (except digit 1 and 7 where having the blockwise skip connections helped). Using global max pooling instead of global average pooling led to worse detection performance for all digits. For all digits the optimization was better with the proposed architecture. Removing skip connections or using global max pooling made the optimization take longer to converge, made loss curves not as smooth and made the loss converged to a higher value. The corresponding FROC curves, FAUC barplot, and FAUC, FPavg, FNavg and Sensitivity Tables are given in \ref{appendix:resultsMNISTarch}.

\subsection{Detection of brain lesions}
\label{sec:baseline}
In the brain dataset, we compare the performance of the weakly supervised methods for the detection enlarged perivascular spaces (PVS) by evaluating them on the left-out test set of 1000 scans, and in four brain regions: midbrain, hippocampi, basal ganglia, and centrum semiovale. 

\begin{figure*}[!t]
\centering
\includegraphics[height=9cm]{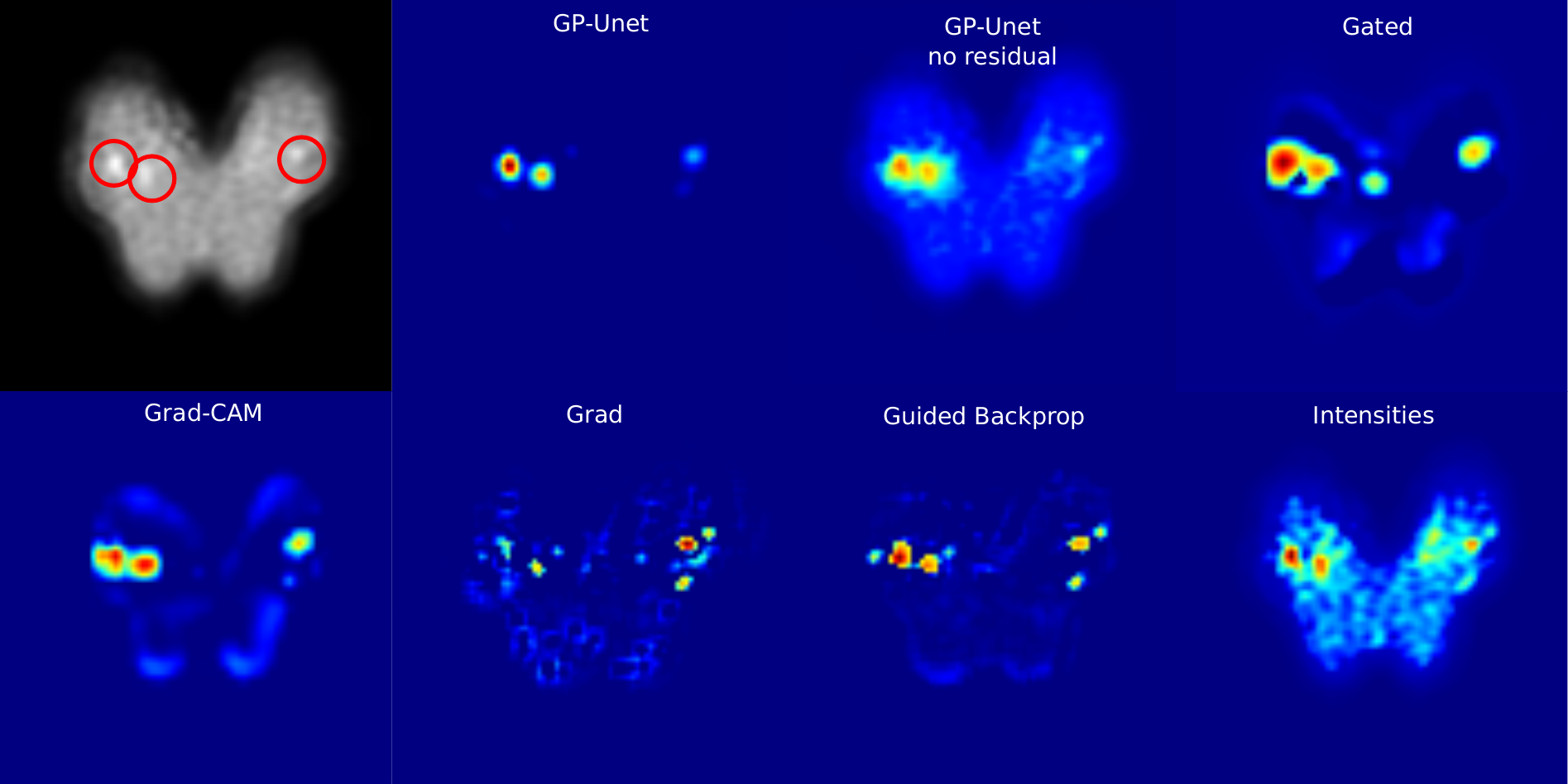}
\caption{\textbf{Attentions maps in the midbrain.} The top left image shows the slice of an example image of the midbrain after preprocessing, with PVS indicated with red circles. The other images correspond to attention maps computed for that same slice. Red values correspond to high values in the attention maps. The intensity baseline method in the bottom right corner is actually the same as the image in the upper left corner but with a different color map. Values in attention maps are not bounded, and the maximum varies between images and methods. For the visualization, we chose the scaling of attention maps to best show the range of values in each image.}
\label{fig:attMapMid}
\end{figure*}

\begin{figure*}[!t]
\centering
\includegraphics[height=7cm]{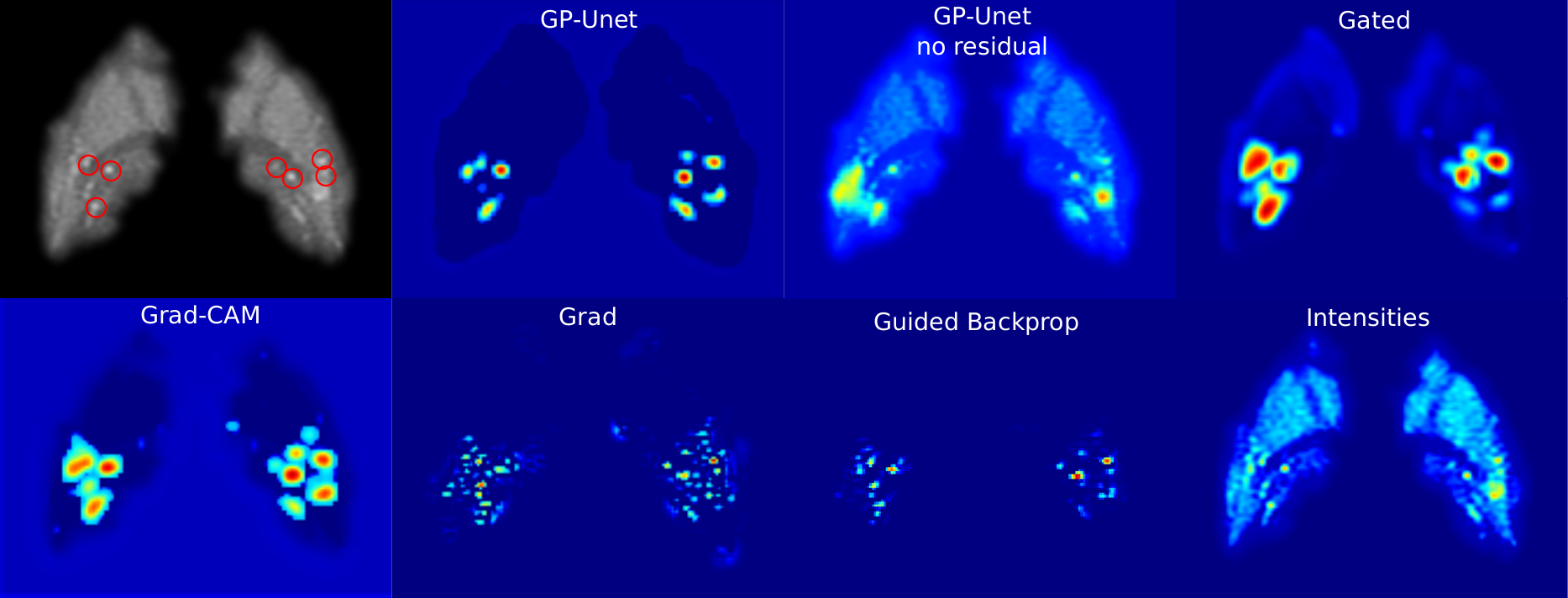}
\caption{\textbf{Attentions maps in the basal ganglia.}}
\label{fig:attMapBG}
\end{figure*}

\begin{figure*}[!t]
\centering
\includegraphics[height=10.5cm]{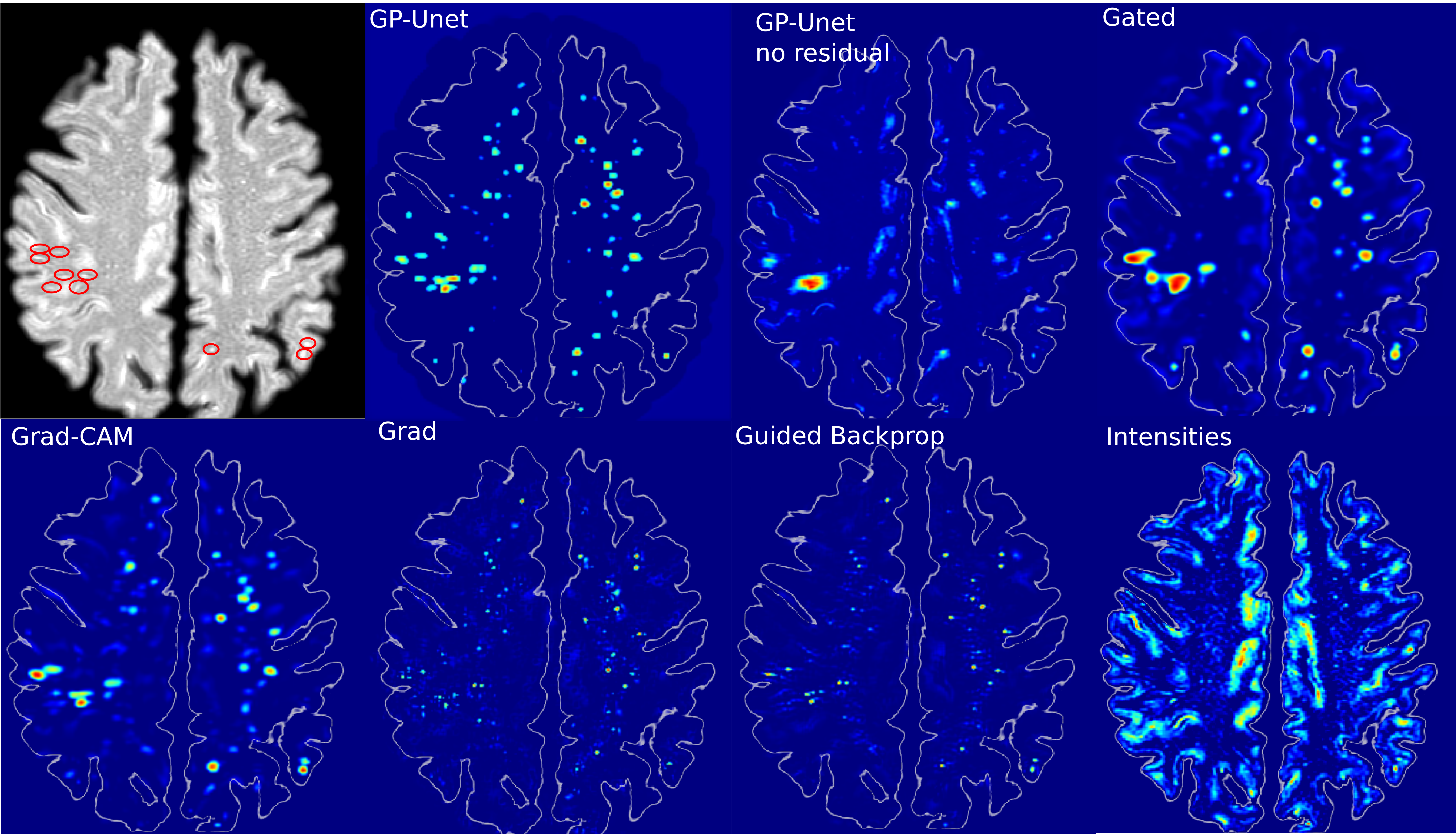}
\caption{\textbf{Attentions maps in the centrum semiovale.} Contours of the brain have been delineated in white for better visualization.}
\label{fig:attMapCSO}
\end{figure*}

Figures \ref{fig:attMapMid} - \ref{fig:attMapCSO} show attention maps for all methods in the four regions.
Figure \ref{fig:froc} shows FROC curves for all methods in the brain datasets. Table \ref{table:fauc} shows the corresponding FAUCs. Table \ref{table:sensitivity} and \ref{table:fpavg} show the sensitivity and FPavg measured at the operating point chosen for each method as described in Section \ref{sec:postproc}.

Judging from Tables \ref{table:fauc}, \ref{table:sensitivity} and \ref{table:fpavg}, the methods achieving the best results are GP-Unet, Grad-CAM and Guid-backpropagation. Unlike the results on MNIST datasets, there is no method consistently better than others for all regions. In the midbrain and basal ganglia, Guided-backpropagation reaches the best results of all methods, and in all three metrics, with the exception of FPavg in the basal ganglia. In the hippocampi, GP-Unet reaches the best results of all methods, and in all four metrics. In the centrum semiovale, GP-Unet and Grad-CAM achieve the best results, and have a similar performance.
Intensity thresholding reaches a competitive performance in the midbrain and basal ganglia, but completely fails in the hippocampi and centrum semiovale because it highlights many false positives, corresponding to other hyperintense structures. Surrounding cerebrospinal fluid, white matter hyperintensities, and sulci are examples of these structures.

In Figure \ref{fig:froc}, the sensitivity and FPavg between two series of annotations of the same scans from the same rater (green triangle) gives an idea of the difficulty of detecting PVS in each region. In the midbrain and hippocampi, PVS are relatively easy to identify, as they are the only hyperintense lesions visible on T2 images. On the contrary, the detection of PVS in the basal ganglia and centrum semiovale is much more challenging, because in those regions other hyperintense structures that look similar to enlarged perivascular spaces. In all regions, the performance of the automated methods come close to the intra-rater agreement. This intrarater agreement was however computed on a substantially smaller set -- 40 vs 1000 scans -- and shorter annotation period -- 1 week vs several months. Interestingly, several methods highlight the same false positives. After visual checking by experts, many of these false positives appear to be PVS annotated by the rater.
In the set of 40 scans used the the intrarater measures, 68 percent of false positive detections of GP-Unet in the centrum semiovale were PVS. More precisely, 39 percent of false positives were enlarged PVS and 29 percent were slightly enlarged PVS.

\begin{figure*}[!t]
\centering
\includegraphics[height=7cm]{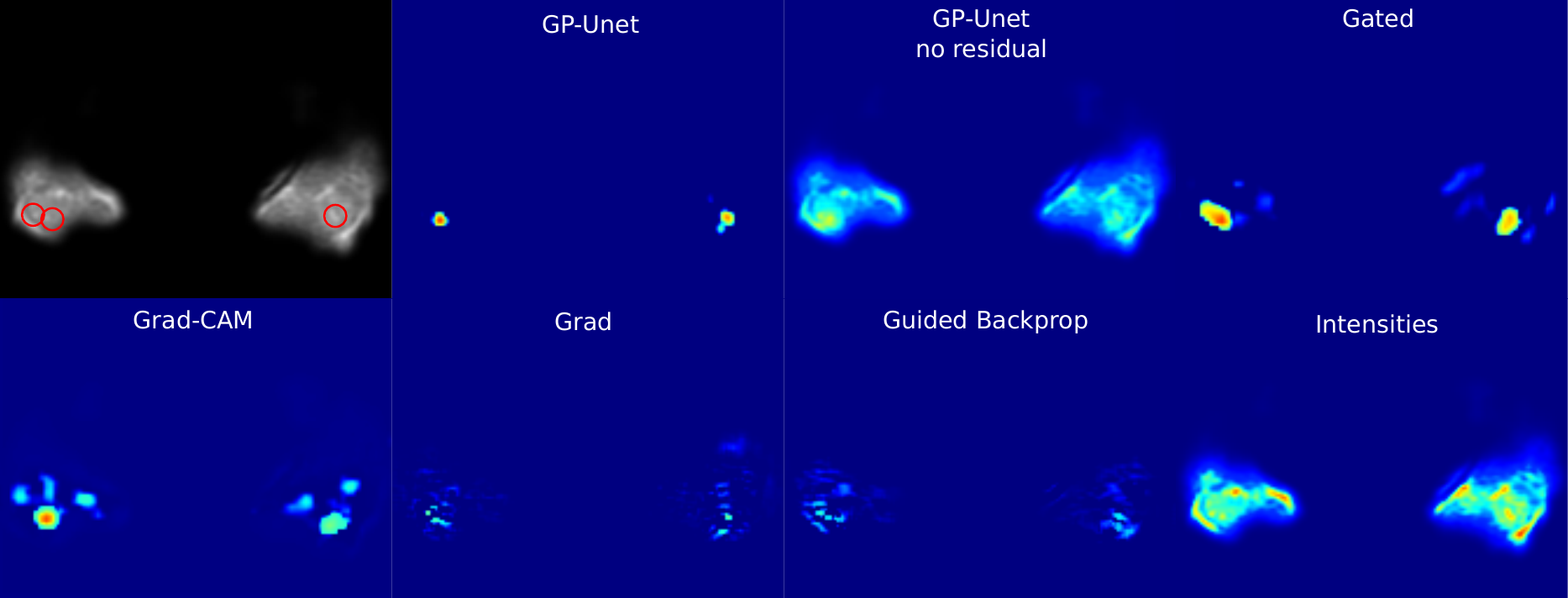}
\caption{\textbf{Attentions maps in the hippocampi.}}
\label{fig:attMapHip}
\end{figure*}

%------------------------------------------------

\section{Discussion}

Overall, results showed that weakly supervised methods can detect PVS almost as well as expert raters. The performance of the best detection methods was close to the intrarater agreement. The interrater agreement is also probably lower than this intrarater agreement. Finally, further visual inspection also revealed that many of the false positives correspond to PVS that were not annotated by the human rater. We especially noticed that annotating all PVS was difficult for the expert rater in scans with many PVS.

We compared six weakly supervised detection methods in two datasets. We showed that the proposed method could be used with either 2D or 3D networks. For all methods, 2D networks in the MNIST datasets converged substantially faster (hours) than the 3D networks in the brain dataset (days). In MNIST datasets for regression, GP-Unet no residual \citep{dubost2017} and GP-Unet (this article) perform significantly better than all other methods, probably because they can combine the information of different scales more effectively than other methods. For GP-Unet no residual, part of this performance difference can also be explained by the larger number of parameters and larger receptive field (Section \ref{sec:arch}). On the contrary, for GP-Unet, the number of parameters is comparable to that of the other methods. 
In the brain dataset, the best methods are Guided-backpropagation \citep{springenberg2014} with 74.1 average FAUC over regions, GP-Unet with 72.0 average FAUC, and Grad-CAM \cite{selvaraju2017} with 70.5 average FAUC. As GP-Unet performs either similarly to or better than Grad-CAM depending on the region, given a new weakly supervised detection task, we would consequently recommend Guided-backpropagation and GP-Unet.

Grad-CAM and GP-Unet reach similar FAUCs (Table \ref{table:fauc}) in the basal ganglia and centrum semiovale. However, GP-Unet outperforms Grad-CAM in the midbrain and by a large margin in the hippocampi. In these two regions, at the operating point Grad-CAM suffers from more false positives than GP-Unet, while having a similar or worse sensitivity (Table \ref{table:fpavg} and \ref{table:sensitivity}). The attention maps of the hippocampi (Figure \ref{fig:attMapHip}) -- and to some extent those of the midbrain (Figure \ref{fig:attMapMid}) -- show that GP-Unet is less distracted by the surrounding cerebrospinal fluid than Grad-CAM -- or the methods emphasizing intensities (GP-Unet no residual, Intensities). The attention maps of Grad-CAM and GP-Unet share most of the false positive detections. Most of these false positives are PVS that were not annotated by the rater. Overall, the attention maps of GP-Unet are also sharper than the ones of Grad-CAM, probably because GP-Unet can compute attention maps at a higher resolution: the resolution of the input image. 

The motivation of Gated Attention \citep{schlemper2018} is similar to that of GP-Unet: combining multiscale information in the computation of attention maps. In the MNIST datasets, while Gated Attention and GP-Unet reach a similar detection performance when optimized with classification objectives, contrary to GP-Unet, Gated Attention rarely benefits from the regression objective. More generally, Gated Attention seems to benefit less often from the regression objective than the other methods. These results suggest that gate mechanisms may harm the detection performance for networks optimized with regression objectives, and that a simple concatenation of feature maps should be preferred. In the brain datasets, Gated Attention works better than the intensity baseline, Grad \citep{simonyan2014}, and GP-Unet no residual, but performs significantly worse than Grad-CAM, Guided-backpropagation, and GP-Unet. One should also keep in mind that Gated Attention was originally proposed for deeper networks. In case of shallow networks, this method may not reach its full potential, as it benefits only from few (two on our case) different feature scales. 

We mentioned above that the attention maps of GP-Unet are sharper than those of Grad-CAM. In \ref{appendix:resultsMNISTarch}, we investigate the influence of the architecture and compare attention maps of GP-Unet, GP-Unet without blockwise skip connections (GP-Unet No Skip) and GP-Unet with global max pooling instead of global average pooling (GP-Unet Max Pool). Removing the skip connections does not seem to make the attention less compact. Using global max pooling does make the attention maps more compact but increases the number of false negatives. GP-Unet may have more compact attention maps than Grad-CAM on the basic architecture thanks to the upsampling path in GP-Unet. To compute the attention at full input resolution with Grad-CAM, the attention maps need to be interpolated, resulting in les compact attention maps. GP-Unet may have more compact attention maps than Gated Attention because concatenating feature maps might be more efficient (maybe easier to optimize) in combining multiscale features than using the gated attention.

Due to the special properties of the PVS detection problem in the brain datasets, intensity thresholding provides a simple approach to solving the same problem. Although intensity thresholding yields the worst results in hippocampi, basal ganglia, and centrum semiovale, it achieves the second best FAUC in the midbrain. This high performance results from the effective region masking specific to the midbrain: because PVS are almost always in the center of this region, we can erode the border of the region mask, and eliminate the hyperintense cerebrospinal fluid surrounding the midbrain. As there are no other visible lesions in the midbrain, all remaining hyperintensities correspond to PVS.

In the datasets where the intensity method achieved good or reasonable results (midbrain and basal ganglia), Guided-backpropagation performed best. In the datasets where the intensity method failed (hippocampi and centrum semiovale), GP-Unet reached the best performance (similar to that of Grad-CAM in the centrum semiovale). More generally, gradients methods seem to work best when the target objects are also the most salient objects, while CAM methods work best when saliency alone is not discriminative enough. This observation can also be extended to the MNIST datasets, where saliency alone is not sufficient, and regression CAM methods (Gated Attention excluded) outperform regression gradient methods.

Recently \cite{adebayo2018} showed that, for Guided-backpropagation, classification networks trained with random labels obtained similar attention maps as networks trained with the correct labels, hinting that attention maps method may focus more on salient objects in the image than the target object. In these experiments, attention maps computed with Grad and Grad-CAM obtained better results.
Adebayo et al. warn of the evaluation of attention maps by only visual appeal, and advocate more rigorous forms of evaluation. This fits exactly with the purpose of the current article, in which we aimed to quantify the detection performance of attention maps in large real world datasets. 

For the evaluation of the detection of PVS, images were annotated by a single rater. With the same resources, we could also have had multiple raters annotating fewer scans and use their consensus for the evaluation, which may reduce the risk of mislabeling. We preferred to evaluate the detection using more scans to better encompass the anatomical variability, and we quantified the performance of the single rater by computing her intra-rater agreement on a smaller set.

In our preliminary work on PVS detection in the basal ganglia using GP-Unet no residual \citep{dubost2017} we obtained slightly different results than what is presented in the current work. This reflects differences in the test data set, the annotations, method and postprocessing. Our previous annotations \citep{dubost2017} were done directly on the segmented and cropped basal ganglia, while the annotations of the current work were done on the full scan. The rater sometimes annotated lesions at the borders of the basal ganglia which are barely visible after preprocessing. In addition, the current work also includes scans without annotations (because the rater found no lesion), where there could have been errors in finding the slice evaluated by the rater.
In the current work, Grad reaches better results than in \cite{dubost2017}, because it benefits from the more sophisticated postprocessing: the non-maximum suppression clears the noise in the attention maps.

Next to the methods presented in this paper, we experimented with the perturbation method with masks proposed by \cite{petsiuk2018}. For this method, masks are first sampled in a low dimensional space and resized to the size of the input image. It appeared that the size of this lower dimensional needs to be adapted to the size of the target object in the image. If the target objects are small, one may need to sample relatively large masks. In our experiments, we experimented with a range of values for the size of this low dimensional space, and did not manage to compute discriminative attention maps for PVS, that are small objects relatively to the image resolution.

The work presented in this article implies that pixel-level annotations may not be needed to train accurate models for detection problems. This is especially relevant in medical imaging, where annotation requires expert knowledge and high quality annotations are therefore difficult to obtain. Weakly supervised methods enable learning from large databases, such as UK biobank \citep{sudlow2015} or Framingham study \citep{maillard2016}, with less annotation effort, and could also help to reduce the dependence on annotator biases. The global label may even be more reliable, because for some abnormalities raters can agree well on the presence or global burden of the abnormalities but poorly on their boundaries or spatial distribution.

The variety of challenges present in the brain datasets are well suited to the evaluation of weakly-supervised detection methods. Observations and results might generalize to the detection of other types of small objects, such as microinfarcts, microbleeds, or small white matter hyperintensities.

\section{Conclusion}
We proposed a new weakly supervised detection method, GP-Unet, that uses an encoder-decoder architecture optimized only with global labels such as the count of lesions in a brain region. The decoder part upsamples feature maps and enables the computation of attention maps at the resolution of the input image, which thus helps the detection of small objects. We also showed the advantage of using regression objectives over classification objectives for the optimization of weakly supervised detection methods, when the target object appears multiple times in the image.
We compared the proposed method to four state-of-the-art methods on the detection of digits in MNIST-based datasets, and on the detection of enlarged perivascular spaces -- a type of brain lesion -- from 3D brain MRI. The best weakly supervised detection methods were Guided-backpropagation \citep{springenberg2014}, and the proposed method GP-Unet. We noticed that methods based on the gradient of the output of the network, such as Guided-backpropagation, worked best in datasets where the target objects are also the most salient objects. In other datasets, methods using class activation maps, such as GP-Unet, worked best. 
The performance of the detection enlarged perivascular spaces using the weakly supervised methods was close to the intrarater agreement of an expert rater. The proposed method could consequently facilitate studies of enlarged perivascular and help advance research in their etiology and relationship with cerebrovascular diseases.

\section{Acknowledgements}

This research was funded by The Netherlands Organisation for Health Research and Development (ZonMw) Project 104003005, with additional support of Netherlands Organisation for Scientific Research (NWO), project NWO-EW VIDI 639.022.010 and project NWO-TTW Perspectief Programme P15-26. This work was partly carried out on the Dutch national e-infrastructure with the support of SURF Cooperative.

\bibliography{biblio}

\cleardoublepage
\onecolumn
\appendix

\section{\textbf{Results MNIST -- Regression Objectives.}}
\label{appendix:resultsMNISTreg}

\begin{figure}[!htb]
\centering
\includegraphics[width=\textwidth]{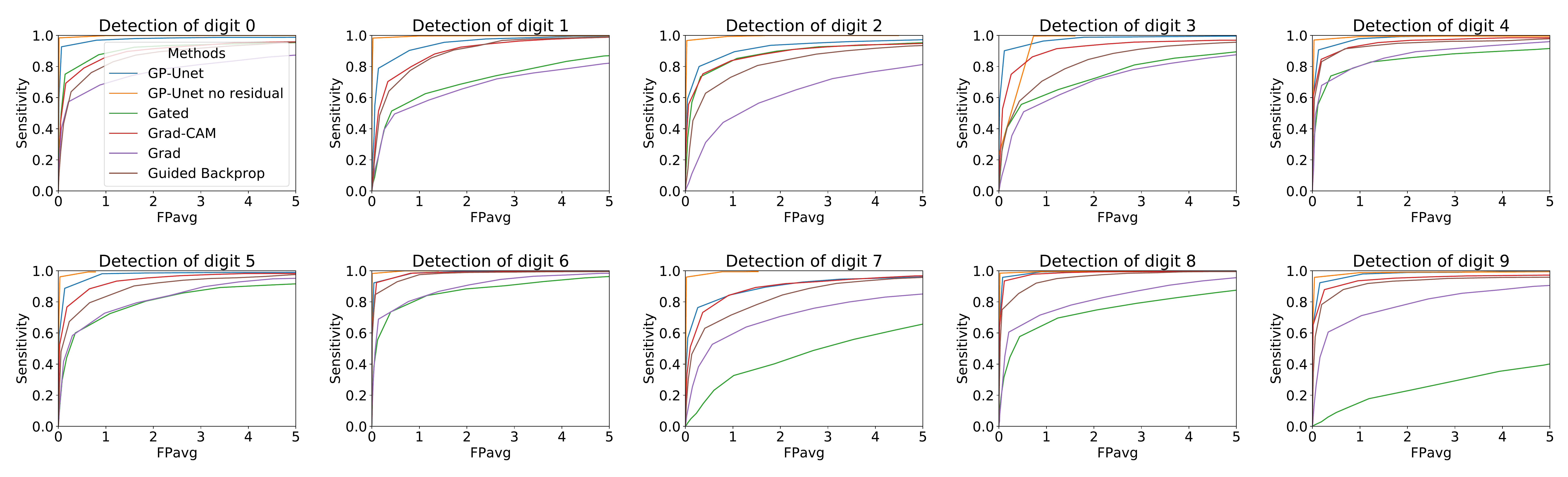}
\caption{\textbf{FROC MNIST regression.} Each subplot corresponds to the detection of a different digit.}
\label{fig:attMapHip}
\end{figure}

\begin{table}[!htb]
\caption{\textbf{FAUCs MNIST regression.} Each row corresponds to the detection of a different digit. 95 percent confidence interval is indicated in brackets. The average and standard deviation of the performance of each method across all digits is given in the last row. Best performance are indicated in bold.}
\resizebox{\textwidth}{!}{%
\begin{tabular}{ccccccc}
\hline
& \begin{tabular}[c]{@{}c@{}}GP-Unet\\ (this paper)\end{tabular} & \begin{tabular}[c]{@{}c@{}}GP-Unet no residual\\ \cite{dubost2017}\end{tabular} & \begin{tabular}[c]{@{}c@{}}Gated Attention\\ \cite{schlemper2018}\end{tabular} & \begin{tabular}[c]{@{}c@{}}Grad-CAM\\ \cite{selvaraju2017}\end{tabular} & \begin{tabular}[c]{@{}c@{}}Grad\\ \cite{simonyan2014}\end{tabular} & \begin{tabular}[c]{@{}c@{}}Guided-backprop\\ \cite{springenberg2014}\end{tabular} \\ \hline
0   & 97.4 (96.4 - 98.4) & \textbf{99.7 (99.5 - 99.8)} & 90.5 (88.9 - 92.0) & 89.2 (87.4 - 90.9) & 76.1 (73.8 - 78.3) & 85.8 (83.8 - 87.5) \\
1   & 94.4 (93.5 - 95.1) & \textbf{99.5 (99.3 - 99.7)} & 69.7 (67.4 - 72.0) & 89.1 (87.8 - 90.4) & 66.3 (64.0 - 68.5) & 88.3 (86.9 - 89.5) \\
2   & 91.7 (90.5 - 92.8) & \textbf{99.3 (99.0 - 99.5)} & 87.5 (85.8 - 89.3) & 87.8 (85.8 - 89.6) & 61.0 (58.5 - 63.6) & 81.3 (79.0 - 83.3) \\
3   & 97.3 (96.6 - 97.9) & \textbf{99.6 (99.2 - 99.9)} & 73.4 (70.9 - 75.6) & 91.2 (89.7 - 92.6) & 69.8 (67.4 - 72.0) & 82.1 (80.0 - 84.1) \\
4   & 97.8 (97.2 - 98.3) & \textbf{99.0 (98.5 - 99.5)} & 83.9 (81.8 - 86.0) & 95.0 (93.8 - 96.0) & 86.4 (84.8 - 87.9) & 93.5 (92.1 - 94.6) \\
5   & 97.1 (96.3 - 97.8) & \textbf{98.9 (98.4 - 99.4)} & 79.6 (77.4 - 81.8) & 93.6 (92.4 - 94.7) & 81.5 (79.7 - 83.3) & 88.9 (87.2 - 90.5) \\
6   & 98.6 (98.2 - 99.0) & \textbf{99.9 (99.8 - 99.9)} & 86.5 (84.4 - 88.5) & 98.4 (97.8 - 98.8) & 89.3 (87.9 - 90.6) & 97.3 (96.6 - 97.9) \\
7   & 89.3 (87.6 - 91.0) & \textbf{99.1 (98.5 - 99.6)} & 43.9 (41.2 - 46.5) & 88.5 (86.9 - 90.0) & 69.6 (67.3 - 71.9) & 82.4 (80.7 - 84.0) \\
8   & \textbf{98.8 (98.3 - 99.2)} & \textbf{99.5 (99.1 - 99.8)} & 73.6 (70.8 - 76.2) & 98.2 (97.7 - 98.7) & 81.3 (79.1 - 83.3) & 95.2 (94.5 - 96.0) \\
9   & \textbf{97.6 (96.8 - 98.3)} & \textbf{98.6 (98.1 - 99.1)} & 24.8 (22.6 - 27.1) & 94.3 (93.0 - 95.5) & 77.9 (75.8 - 80.0) & 91.8 (90.4 - 93.2) \\
\hdashline[1.5pt/5pt]
Average & 96.0 +/- 3.0        & 99.3 +/- 0.4        & 71.3 +/- 20.0       & 92.5 +/- 3.7        & 75.9 +/- 8.6        & 88.7 +/- 5.4        \\ \hline
\end{tabular}
}
\end{table}

\begin{table}[!htb]
\caption{\textbf{Sensitivity MNIST regression.} Each row corresponds to the detection of a different digit. 95 percent confidence interval is indicated in brackets. The average and standard deviation of the performance of each method across all digits is given in the last row. Best performance are indicated in bold.}
\resizebox{\textwidth}{!}{%
\begin{tabular}{ccccccc}
\hline
& \begin{tabular}[c]{@{}c@{}}GP-Unet\\ (this paper)\end{tabular} & \begin{tabular}[c]{@{}c@{}}GP-Unet no residual\\ \cite{dubost2017}\end{tabular} & \begin{tabular}[c]{@{}c@{}}Gated Attention\\ \cite{schlemper2018}\end{tabular} & \begin{tabular}[c]{@{}c@{}}Grad-CAM\\ \cite{selvaraju2017}\end{tabular} & \begin{tabular}[c]{@{}c@{}}Grad\\ \cite{simonyan2014}\end{tabular} & \begin{tabular}[c]{@{}c@{}}Guided-backprop\\ \cite{springenberg2014}\end{tabular} \\ \hline
0   & 92.7 (91.4 - 93.9) & \textbf{98.4 (97.8 - 99.1)} & 75.1 (73.1 - 77.0) & 69.2 (67.2 - 71.3) & 57.3 (55.3 - 59.2) & 63.8 (61.7 - 65.8) \\
1   & 78.9 (77.3 - 80.4) & \textbf{98.3 (97.8 - 98.8)} & 51.4 (49.4 - 53.4) & 70.3 (68.6 - 72.1) & 49.5 (47.7 - 51.2) & 64.2 (62.5 - 65.8) \\
2   & 80.0 (78.2 - 81.8) & \textbf{96.7 (95.9 - 97.5)} & 73.5 (71.5 - 75.5) & 75.3 (73.4 - 77.2) & 44.0 (41.9 - 46.1) & 62.8 (60.9 - 64.7) \\
3   & 90.1 (88.8 - 91.5) & \textbf{97.9 (97.4 - 98.4)} & 55.6 (53.5 - 57.7) & 75.0 (73.1 - 76.9) & 50.9 (48.9 - 52.9) & 57.7 (55.5 - 59.9) \\
4   & 90.7 (89.3 - 92.1) & \textbf{97.0 (96.3 - 97.8)} & 73.9 (71.8 - 76.0) & 84.6 (83.0 - 86.3) & 67.9 (66.0 - 69.7) & 83.1 (81.4 - 84.8) \\
5   & 88.7 (87.2 - 90.2) & \textbf{96.1 (95.2 - 97.0)} & 60.0 (57.8 - 62.2) & 76.7 (74.6 - 78.7) & 58.4 (56.2 - 60.5) & 67.3 (65.2 - 69.4) \\
6   & 92.2 (91.0 - 93.5) & \textbf{98.3 (97.7 - 98.9)} & 73.6 (71.5 - 75.6) & 92.4 (91.1 - 93.7) & 68.9 (67.1 - 70.7) & 84.6 (82.9 - 86.2) \\
7   & 76.3 (74.6 - 78.1) & \textbf{95.9 (94.9 - 97.0)} & 32.7 (30.7 - 34.6) & 73.2 (71.4 - 75.0) & 52.7 (50.7 - 54.6) & 63.0 (61.2 - 64.8) \\
8   & 95.8 (95.0 - 96.5) & \textbf{98.5 (98.0 - 98.9)} & 57.7 (55.5 - 59.9) & 93.5 (92.5 - 94.4) & 60.5 (58.7 - 62.4) & 75.0 (73.3 - 76.7) \\
9   & 92.3 (91.1 - 93.5) & \textbf{95.8 (95.0 - 96.6)} & 17.8 (16.2 - 19.3) & 87.9 (86.5 - 89.3) & 60.6 (58.7 - 62.5) & 78.3 (76.6 - 80.0) \\
\hdashline[1.5pt/5pt]
Average & 87.8 +/- 6.4        & 97.3 +/- 1.0        & 57.1 +/- 18.2       & 79.8 +/- 8.6        & 57.1 +/- 7.5        & 70.0 +/- 9.0        \\ \hline
\end{tabular}
}
\end{table}

\begin{table}[!htb]
\caption{\textbf{FPavg MNIST regression.} Each row corresponds to the detection of a different digit. 95 percent confidence interval is indicated in brackets. The average and standard deviation of the performance of each method across all digits is given in the last row. Best performance are indicated in bold.}
\resizebox{\textwidth}{!}{%
\begin{tabular}{ccccccc}
\hline
& \begin{tabular}[c]{@{}c@{}}GP-Unet\\ (this paper)\end{tabular} & \begin{tabular}[c]{@{}c@{}}GP-Unet no residual\\ \cite{dubost2017}\end{tabular} & \begin{tabular}[c]{@{}c@{}}Gated Attention\\ \cite{schlemper2018}\end{tabular} & \begin{tabular}[c]{@{}c@{}}Grad-CAM\\ \cite{selvaraju2017}\end{tabular} & \begin{tabular}[c]{@{}c@{}}Grad\\ \cite{simonyan2014}\end{tabular} & \begin{tabular}[c]{@{}c@{}}Guided-backprop\\ \cite{springenberg2014}\end{tabular} \\ \hline
0   & 0.07 (0.05 - 0.08) & \textbf{0.02 (0.01 - 0.03)} & 0.14 (0.11 - 0.17) & 0.16 (0.13 - 0.18) & 0.21 (0.18 - 0.24) & 0.27 (0.23 - 0.31) \\
1   & 0.14 (0.11 - 0.16) & \textbf{0.02 (0.01 - 0.03)} & 0.42 (0.36 - 0.47) & 0.34 (0.30 - 0.38) & 0.48 (0.42 - 0.53) & 0.36 (0.31 - 0.40) \\
2   & 0.29 (0.25 - 0.32) & \textbf{0.03 (0.01 - 0.04)} & 0.32 (0.28 - 0.36) & 0.37 (0.33 - 0.41) & 0.79 (0.73 - 0.86) & 0.42 (0.38 - 0.47) \\
3   & 0.11 (0.09 - 0.14) & \textbf{0.00 (0.00 - 0.00)} & 0.47 (0.42 - 0.52) & 0.26 (0.22 - 0.29) & 0.52 (0.46 - 0.58) & 0.43 (0.38 - 0.48) \\
4   & 0.13 (0.10 - 0.15) & \textbf{0.03 (0.02 - 0.04)} & 0.39 (0.34 - 0.43) & 0.18 (0.15 - 0.21) & 0.19 (0.16 - 0.22) & 0.19 (0.16 - 0.22) \\
5   & 0.13 (0.11 - 0.16) & \textbf{0.03 (0.02 - 0.05)} & 0.36 (0.31 - 0.40) & 0.18 (0.15 - 0.21) & 0.30 (0.25 - 0.34) & 0.23 (0.20 - 0.26) \\
6   & 0.04 (0.03 - 0.06) & \textbf{0.01 (0.00 - 0.01)} & 0.40 (0.35 - 0.44) & 0.10 (0.08 - 0.12) & 0.14 (0.11 - 0.17) & 0.06 (0.05 - 0.08) \\
7   & 0.26 (0.23 - 0.29) & \textbf{0.02 (0.01 - 0.03)} & 1.02 (0.93 - 1.11) & 0.36 (0.32 - 0.41) & 0.57 (0.51 - 0.62) & 0.41 (0.36 - 0.45) \\
8   & 0.08 (0.06 - 0.10) & \textbf{0.02 (0.01 - 0.03)} & 0.44 (0.38 - 0.49) & 0.11 (0.09 - 0.14) & 0.21 (0.17 - 0.25) & 0.07 (0.05 - 0.09) \\
9   & 0.16 (0.13 - 0.18) & \textbf{0.04 (0.03 - 0.06)} & 1.19 (1.07 - 1.30) & 0.25 (0.22 - 0.28) & 0.33 (0.29 - 0.38) & 0.19 (0.16 - 0.23) \\
\hdashline[1.5pt/5pt]
Average & 0.14 +/- 0.08      & 0.02 +/- 0.01      & 0.51 +/- 0.31      & 0.23 +/- 0.10      & 0.37 +/- 0.20      & 0.26 +/- 0.13      \\ \hline
\end{tabular}
}
\end{table}

\begin{table}[!htb]
\caption{\textbf{FNavg MNIST regression.} Each row corresponds to the detection of a different digit. 95 percent confidence interval is indicated in brackets. The average and standard deviation of the performance of each method across all digits is given in the last row. Best performance are indicated in bold.}
\resizebox{\textwidth}{!}{%
\begin{tabular}{ccccccc}
\hline
& \begin{tabular}[c]{@{}c@{}}GP-Unet\\ (this paper)\end{tabular} & \begin{tabular}[c]{@{}c@{}}GP-Unet no residual\\ \cite{dubost2017}\end{tabular} & \begin{tabular}[c]{@{}c@{}}Gated Attention\\ \cite{schlemper2018}\end{tabular} & \begin{tabular}[c]{@{}c@{}}Grad-CAM\\ \cite{selvaraju2017}\end{tabular} & \begin{tabular}[c]{@{}c@{}}Grad\\ \cite{simonyan2014}\end{tabular} & \begin{tabular}[c]{@{}c@{}}Guided-backprop\\ \cite{springenberg2014}\end{tabular} \\ \hline
0   & 0.11 (0.08 - 0.14) & \textbf{0.02 (0.01 - 0.03)} & 0.37 (0.32 - 0.43) & 0.47 (0.41 - 0.53) & 0.62 (0.55 - 0.69) & 0.54 (0.47 - 0.61) \\
1   & 0.35 (0.30 - 0.40) & \textbf{0.03 (0.02 - 0.05)} & 0.75 (0.67 - 0.82) & 0.49 (0.43 - 0.56) & 0.82 (0.73 - 0.91) & 0.58 (0.52 - 0.64) \\
2   & 0.30 (0.25 - 0.34) & \textbf{0.05 (0.03 - 0.07)} & 0.40 (0.34 - 0.45) & 0.37 (0.32 - 0.43) & 0.82 (0.73 - 0.90) & 0.56 (0.50 - 0.63) \\
3   & 0.15 (0.12 - 0.18) & \textbf{0.04 (0.02 - 0.05)} & 0.58 (0.52 - 0.65) & 0.34 (0.30 - 0.39) & 0.68 (0.61 - 0.76) & 0.56 (0.49 - 0.62) \\
4   & 0.12 (0.10 - 0.15) & \textbf{0.05 (0.03 - 0.06)} & 0.36 (0.31 - 0.41) & 0.22 (0.18 - 0.25) & 0.49 (0.43 - 0.56) & 0.25 (0.21 - 0.29) \\
5   & 0.14 (0.11 - 0.16) & \textbf{0.05 (0.04 - 0.07)} & 0.50 (0.44 - 0.56) & 0.29 (0.25 - 0.34) & 0.54 (0.48 - 0.60) & 0.42 (0.37 - 0.47) \\
6   & 0.11 (0.08 - 0.13) & \textbf{0.02 (0.01 - 0.03)} & 0.35 (0.31 - 0.40) & 0.09 (0.07 - 0.12) & 0.47 (0.41 - 0.52) & 0.22 (0.18 - 0.26) \\
7   & 0.36 (0.31 - 0.41) & \textbf{0.04 (0.03 - 0.06)} & 1.00 (0.90 - 1.10) & 0.43 (0.37 - 0.49) & 0.73 (0.65 - 0.81) & 0.58 (0.51 - 0.65) \\
8   & 0.07 (0.05 - 0.09) & \textbf{0.02 (0.01 - 0.04)} & 0.58 (0.50 - 0.65) & 0.11 (0.08 - 0.13) & 0.59 (0.52 - 0.66) & 0.37 (0.32 - 0.43) \\
9   & 0.11 (0.09 - 0.14) & \textbf{0.07 (0.05 - 0.09)} & 1.22 (1.09 - 1.34) & 0.20 (0.16 - 0.24) & 0.63 (0.55 - 0.70) & 0.36 (0.31 - 0.41) \\
\hdashline[1.5pt/5pt]
Average & 0.18 +/- 0.10      & 0.04 +/- 0.02      & 0.61 +/- 0.28      & 0.30 +/- 0.14      & 0.64 +/- 0.12      & 0.44 +/- 0.13      \\ \hline
\end{tabular}
}
\end{table}

\cleardoublepage
\section{\textbf{Results MNIST -- Classification Objectives.}}
\label{appendix:resultsMNISTclassification}

\begin{figure*}[!htb]
\centering
\includegraphics[width=\textwidth]{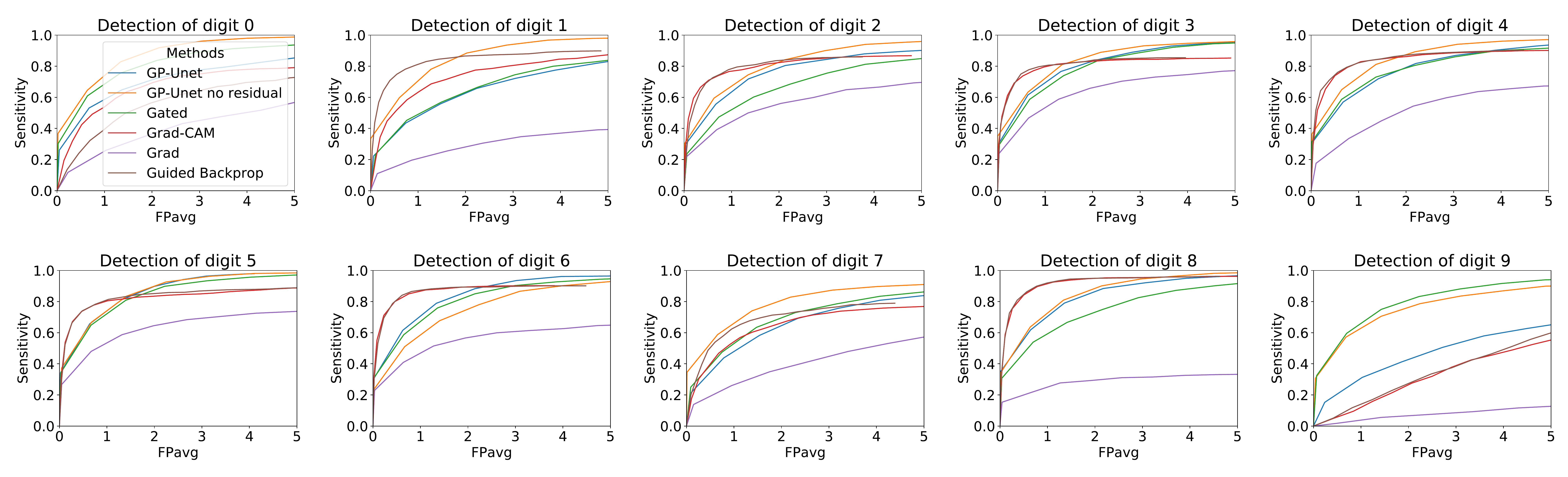}
\caption{\textbf{FROC MNIST classification.} Each subplot corresponds to the detection of a different digit.}
\label{fig:attMapHip}
\end{figure*}

\begin{table}[!htb]
\caption{\textbf{FAUCs MNIST classification.} Each row corresponds to the detection of a different digit. 95 percent confidence interval is indicated in brackets. The average and standard deviation of the performance of each method across all digits is given in the last row. Best performance are indicated in bold.}
\resizebox{\textwidth}{!}{%
\begin{tabular}{ccccccc}
\hline
& \begin{tabular}[c]{@{}c@{}}GP-Unet\\ (this paper)\end{tabular} & \begin{tabular}[c]{@{}c@{}}GP-Unet no residual\\ \cite{dubost2017}\end{tabular} & \begin{tabular}[c]{@{}c@{}}Gated Attention\\ \cite{schlemper2018}\end{tabular} & \begin{tabular}[c]{@{}c@{}}Grad-CAM\\ \cite{selvaraju2017}\end{tabular} & \begin{tabular}[c]{@{}c@{}}Grad\\ \cite{simonyan2014}\end{tabular} & \begin{tabular}[c]{@{}c@{}}Guided-backprop\\ \cite{springenberg2014}\end{tabular} \\ \hline
0   & 69.3 (66.7 - 71.7) & \textbf{85.8 (84.6 - 87.1)} & 79.5 (77.4 - 81.6) & 65.2 (61.8 - 68.6) & 37.5 (34.4 - 40.6) & 54.1 (50.7 - 57.7) \\
1   & 62.6 (60.3 - 65.0) & \textbf{83.9 (82.5 - 85.3)} & 63.8 (61.3 - 66.3) & 72.3 (69.5 - 75.3) & 28.6 (25.6 - 31.7) & \textbf{83.2 (80.3 - 86.1)} \\
2   & 75.5 (73.4 - 77.5) & \textbf{80.3 (78.7 - 81.9)} & 65.9 (63.7 - 68.3) & \textbf{80.3 (76.9 - 83.6)} & 54.9 (50.7 - 58.6) & \textbf{80.1 (76.5 - 83.5)} \\
3   & 80.6 (79.0 - 82.2) & \textbf{83.4 (81.7 - 84.9)} & 78.9 (77.0 - 80.7) & \textbf{80.7 (77.0 - 84.3)} & 63.7 (59.9 - 67.4) & \textbf{80.9 (77.2 - 84.5)} \\
4   & 77.1 (74.9 - 79.1) & \textbf{84.1 (82.6 - 85.7)} & 77.1 (75.2 - 79.0) & \textbf{83.7 (80.7 - 86.5)} & 51.3 (47.4 - 55.0) & \textbf{84.3 (81.1 - 87.3)} \\
5   & \textbf{85.5 (83.9 - 86.8)} & \textbf{85.5 (84.3 - 86.7)} & \textbf{83.4 (81.9 - 84.9)} & \textbf{81.7 (78.5 - 84.7)} & 62.0 (58.3 - 65.8) & \textbf{82.5 (79.0 - 85.8)} \\
6   & 82.8 (81.1 - 84.3) & 74.1 (71.9 - 76.2) & 80.0 (77.8 - 81.9) & \textbf{86.6 (83.7 - 89.3)} & 54.6 (50.4 - 58.9) & \textbf{86.4 (83.6 - 89.2)} \\
7   & 63.7 (61.1 - 66.2) & \textbf{77.9 (75.9 - 79.9)} & 67.7 (65.4 - 70.0) & 63.5 (60.1 - 67.0) & 38.4 (35.8 - 41.2) & 68.4 (64.7 - 72.1) \\
8   & 82.7 (81.0 - 84.2) & 84.7 (83.2 - 86.0) & 73.2 (71.1 - 75.3) & \textbf{91.6 (89.4 - 93.6)} & 28.6 (24.1 - 33.1) & \textbf{91.9 (89.9 - 93.7)} \\
9   & 44.3 (41.5 - 47.3) & 74.3 (72.2 - 76.5) & \textbf{78.1 (76.0 - 80.0)} & 30.4 (27.6 - 33.3) & 7.2 (5.7 - 8.8)    & 31.7 (28.8 - 34.7) \\
\hdashline[1.5pt/5pt]
Average & 72.4 +/- 12.0       & 81.4 +/- 4.3        & 74.8 +/- 6.4        & 73.6 +/- 16.7       & 42.7 +/- 17.0       & 74.4 +/- 17.4       \\ \hline
\end{tabular}
}
\end{table}

\begin{table}[!htb]
\caption{\textbf{Sensitivity MNIST classification.} Each row corresponds to the detection of a different digit. 95 percent confidence interval is indicated in brackets. The average and standard deviation of the performance of each method across all digits is given in the last row. Best performance are indicated in bold.}
\resizebox{\textwidth}{!}{%
\begin{tabular}{ccccccc}
\hline
& \begin{tabular}[c]{@{}c@{}}GP-Unet\\ (this paper)\end{tabular} & \begin{tabular}[c]{@{}c@{}}GP-Unet no residual\\ \cite{dubost2017}\end{tabular} & \begin{tabular}[c]{@{}c@{}}Gated Attention\\ \cite{schlemper2018}\end{tabular} & \begin{tabular}[c]{@{}c@{}}Grad-CAM\\ \cite{selvaraju2017}\end{tabular} & \begin{tabular}[c]{@{}c@{}}Grad\\ \cite{simonyan2014}\end{tabular} & \begin{tabular}[c]{@{}c@{}}Guided-backprop\\ \cite{springenberg2014}\end{tabular} \\ \hline
   & 26.0 (24.7 - 27.4) & \textbf{36.7 (34.9 - 38.5)} & 30.3 (28.7 - 31.9) & 19.4 (18.0 - 20.8) & 11.8 (10.6 - 12.9) & 13.6 (12.3 - 14.9) \\
1   & 22.3 (21.0 - 23.6) & \textbf{33.5 (31.9 - 35.1)} & 23.4 (22.1 - 24.7) & 19.4 (18.2 - 20.6) & 11.0 (9.8 - 12.2)  & 25.9 (24.6 - 27.3) \\
2   & \textbf{30.1 (28.6 - 31.7)} & \textbf{30.4 (28.8 - 31.9)} & 23.7 (22.3 - 25.1) & 27.3 (25.8 - 28.8) & 21.6 (20.0 - 23.3) & 25.0 (23.5 - 26.5) \\
3   & 30.3 (28.8 - 31.7) & \textbf{35.1 (33.5 - 36.8)} & 29.8 (28.4 - 31.2) & 28.2 (26.7 - 29.6) & 23.9 (22.3 - 25.6) & 26.9 (25.5 - 28.3) \\
4   & 30.8 (29.2 - 32.4) & \textbf{36.5 (34.8 - 38.3)} & 31.2 (29.6 - 32.8) & 29.6 (28.1 - 31.1) & 17.6 (15.9 - 19.3) & 30.8 (29.2 - 32.3) \\
5   & 34.1 (32.5 - 35.8) & \textbf{37.4 (35.6 - 39.1)} & 34.4 (32.8 - 36.1) & 31.4 (29.8 - 33.0) & 26.7 (25.0 - 28.4) & 30.1 (28.5 - 31.6) \\
6   & \textbf{30.9 (29.4 - 32.3)} & 23.9 (22.7 - 25.0) & \textbf{31.0 (29.6 - 32.5)} & \textbf{31.3 (29.7 - 32.8)} & 22.9 (21.1 - 24.7) & \textbf{30.8 (29.3 - 32.4)} \\
7   & 21.6 (20.3 - 22.8) & \textbf{34.6 (32.9 - 36.3)} & 25.0 (23.5 - 26.5) & 17.5 (16.2 - 18.8) & 13.9 (12.7 - 15.0) & 19.1 (17.9 - 20.3) \\
8   & \textbf{34.7 (33.1 - 36.4)} & \textbf{34.0 (32.4 - 35.6)} & 30.7 (29.2 - 32.2) & \textbf{33.5 (31.9 - 35.1)} & 15.3 (13.3 - 17.3) & \textbf{33.7 (32.1 - 35.3)} \\
9   & 15.2 (13.9 - 16.6) & \textbf{30.7 (29.2 - 32.2)} & \textbf{32.4 (30.8 - 34.0)} & 4.9 (4.1 - 5.7)    & 1.7 (1.1 - 2.2)    & 5.1 (4.3 - 6.0)    \\
\hdashline[1.5pt/5pt]
Average & 27.6 +/- 5.9        & 33.3 +/- 3.9        & 29.2 +/- 3.6        & 24.2 +/- 8.4        & 16.6 +/- 7.1        & 24.1 +/- 8.5        \\ \hline
\end{tabular}
}
\end{table}

\begin{table}[!htb]
\caption{\textbf{FPavg MNIST classification.} Each row corresponds to the detection of a different digit. 95 percent confidence interval is indicated in brackets. The average and standard deviation of the performance of each method across all digits is given in the last row. Best performance are indicated in bold.}
\resizebox{\textwidth}{!}{%
\begin{tabular}{ccccccc}
\hline
& \begin{tabular}[c]{@{}c@{}}GP-Unet\\ (this paper)\end{tabular} & \begin{tabular}[c]{@{}c@{}}GP-Unet no residual\\ \cite{dubost2017}\end{tabular} & \begin{tabular}[c]{@{}c@{}}Gated Attention\\ \cite{schlemper2018}\end{tabular} & \begin{tabular}[c]{@{}c@{}}Grad-CAM\\ \cite{selvaraju2017}\end{tabular} & \begin{tabular}[c]{@{}c@{}}Grad\\ \cite{simonyan2014}\end{tabular} & \begin{tabular}[c]{@{}c@{}}Guided-backprop\\ \cite{springenberg2014}\end{tabular} \\ \hline
0   & 0.05 (0.03 - 0.06) & \textbf{0.02 (0.01 - 0.03)}  & \textbf{0.02 (0.01 - 0.04)} & 0.15 (0.12 - 0.17) & 0.23 (0.20 - 0.26) & 0.21 (0.18 - 0.24) \\
1   & 0.06 (0.05 - 0.08) & \textbf{0.00 (0.00 - 0.01)} & 0.11 (0.08 - 0.13) & 0.10 (0.08 - 0.13) & 0.14 (0.12 - 0.17) & 0.04 (0.02 - 0.05) \\
2   & 0.04 (0.03 - 0.05) & \textbf{0.01 (0.00 - 0.02)}  & 0.06 (0.04 - 0.08) & \textbf{0.03 (0.01 - 0.04)} & 0.05 (0.03 - 0.06) & 0.05 (0.03 - 0.07) \\
3   & 0.02 (0.01 - 0.04) & \textbf{0.00 (0.00 - 0.01)} & 0.04 (0.03 - 0.05) & 0.02 (0.01 - 0.03) & 0.03 (0.02 - 0.04) & 0.03 (0.02 - 0.04) \\
4   & \textbf{0.02 (0.01 - 0.03)} & \textbf{0.02 (0.01 - 0.03)}  & \textbf{0.01 (0.00 - 0.02)} & 0.04 (0.03 - 0.06) & 0.10 (0.08 - 0.13) & 0.03 (0.02 - 0.05) \\
5   & \textbf{0.02 (0.01 - 0.03)} & 0.05 (0.03 - 0.06)  & \textbf{0.03 (0.02 - 0.04)} & 0.04 (0.03 - 0.05) & 0.05 (0.03 - 0.06) & 0.05 (0.03 - 0.06) \\
6   & \textbf{0.03 (0.01 - 0.04)} & \textbf{0.03 (0.02 - 0.04)}  & \textbf{0.02 (0.01 - 0.03)} & \textbf{0.02 (0.01 - 0.03)} & 0.03 (0.02 - 0.05) & 0.03 (0.02 - 0.04) \\
7   & 0.10 (0.08 - 0.12) & \textbf{0.01 (0.00 - 0.02)}  & 0.09 (0.07 - 0.11) & 0.11 (0.08 - 0.13) & 0.15 (0.12 - 0.18) & 0.08 (0.06 - 0.10) \\
8   & \textbf{0.01 (0.00 - 0.01)} & \textbf{0.01 (0.00 - 0.02)}  & 0.04 (0.02 - 0.05) & 0.04 (0.02 - 0.05) & 0.04 (0.03 - 0.06) & 0.03 (0.02 - 0.04) \\
9   & 0.24 (0.20 - 0.27) & \textbf{0.04 (0.03 - 0.05)}  & 0.07 (0.05 - 0.08) & 0.42 (0.38 - 0.45) & 0.48 (0.44 - 0.51) & 0.42 (0.38 - 0.45) \\
\hdashline[1.5pt/5pt]
Average & 0.06 +/- 0.06      & 0.02 +/- 0.01       & 0.05 +/- 0.03      & 0.10 +/- 0.11      & 0.13 +/- 0.13      & 0.10 +/- 0.12      \\ \hline
\end{tabular}
}
\end{table}

\begin{table}[!htb]
\caption{\textbf{FNavg MNIST classification.} Each row corresponds to the detection of a different digit. 95 percent confidence interval is indicated in brackets. The average and standard deviation of the performance of each method across all digits is given in the last row. Best performance are indicated in bold.}
\resizebox{\textwidth}{!}{%
\begin{tabular}{ccccccc}
\hline
& \begin{tabular}[c]{@{}c@{}}GP-Unet\\ (this paper)\end{tabular} & \begin{tabular}[c]{@{}c@{}}GP-Unet no residual\\ \cite{dubost2017}\end{tabular} & \begin{tabular}[c]{@{}c@{}}Gated Attention\\ \cite{schlemper2018}\end{tabular} & \begin{tabular}[c]{@{}c@{}}Grad-CAM\\ \cite{selvaraju2017}\end{tabular} & \begin{tabular}[c]{@{}c@{}}Grad\\ \cite{simonyan2014}\end{tabular} & \begin{tabular}[c]{@{}c@{}}Guided-backprop\\ \cite{springenberg2014}\end{tabular} \\ \hline
0   & 1.13 (1.01 - 1.25) & \textbf{1.07 (0.95 - 1.19)} & 1.11 (0.99 - 1.23) & 1.22 (1.09 - 1.35) & 1.31 (1.17 - 1.44) & 1.28 (1.15 - 1.41) \\
1   & 1.34 (1.20 - 1.48) & \textbf{1.24 (1.11 - 1.38)} & 1.32 (1.18 - 1.46) & 1.36 (1.22 - 1.50) & 1.52 (1.36 - 1.68) & 1.29 (1.15 - 1.43) \\
2   & \textbf{1.15 (1.02 - 1.28)} & \textbf{1.14 (1.02 - 1.27)} & 1.20 (1.07 - 1.32) & 1.17 (1.04 - 1.30) & 1.27 (1.13 - 1.42) & 1.19 (1.06 - 1.32) \\
3   & 1.05 (0.93 - 1.16) & \textbf{1.02 (0.90 - 1.14)} & 1.05 (0.93 - 1.16) & 1.06 (0.95 - 1.18) & 1.14 (1.02 - 1.27) & 1.08 (0.96 - 1.19) \\
4   & 1.09 (0.97 - 1.21) & \textbf{1.05 (0.93 - 1.17)} & 1.08 (0.96 - 1.21) & 1.09 (0.97 - 1.22) & 1.27 (1.13 - 1.42) & 1.09 (0.97 - 1.21) \\
5   & 0.99 (0.88 - 1.10) & \textbf{0.97 (0.86 - 1.08)} & 0.98 (0.87 - 1.10) & 1.00 (0.89 - 1.12) & 1.09 (0.96 - 1.21) & 1.01 (0.90 - 1.12) \\
6   & \textbf{1.11 (0.99 - 1.23)} & 1.16 (1.04 - 1.28) & \textbf{1.11 (0.99 - 1.23)} & \textbf{1.11 (0.99 - 1.23)} & 1.26 (1.12 - 1.40) & \textbf{1.12 (1.00 - 1.24)} \\
7   & 1.26 (1.13 - 1.40) & \textbf{1.17 (1.04 - 1.31)} & 1.24 (1.11 - 1.38) & 1.32 (1.18 - 1.45) & 1.37 (1.23 - 1.51) & 1.29 (1.16 - 1.43) \\
8   & \textbf{1.06 (0.94 - 1.18)} & \textbf{1.06 (0.94 - 1.19)} & 1.09 (0.96 - 1.21) & \textbf{1.07 (0.94 - 1.19)} & 1.34 (1.18 - 1.49) & \textbf{1.06 (0.94 - 1.19)} \\
9   & 1.32 (1.18 - 1.46) & \textbf{1.15 (1.02 - 1.29)} & \textbf{1.15 (1.01 - 1.28)} & 1.46 (1.31 - 1.61) & 1.52 (1.36 - 1.68) & 1.46 (1.31 - 1.61) \\
\hdashline[1.5pt/5pt]
Average & 1.15 +/- 0.11      & 1.10 +/- 0.08      & 1.13 +/- 0.09      & 1.19 +/- 0.14      & 1.31 +/- 0.13      & 1.19 +/- 0.13      \\ \hline
\end{tabular}
}
\end{table}

\cleardoublepage
\section{\textbf{Results MNIST -- GP-Unet architectures.}}
\label{appendix:resultsMNISTarch}

\begin{figure}[!htb]
\centering
\includegraphics[width=\textwidth]{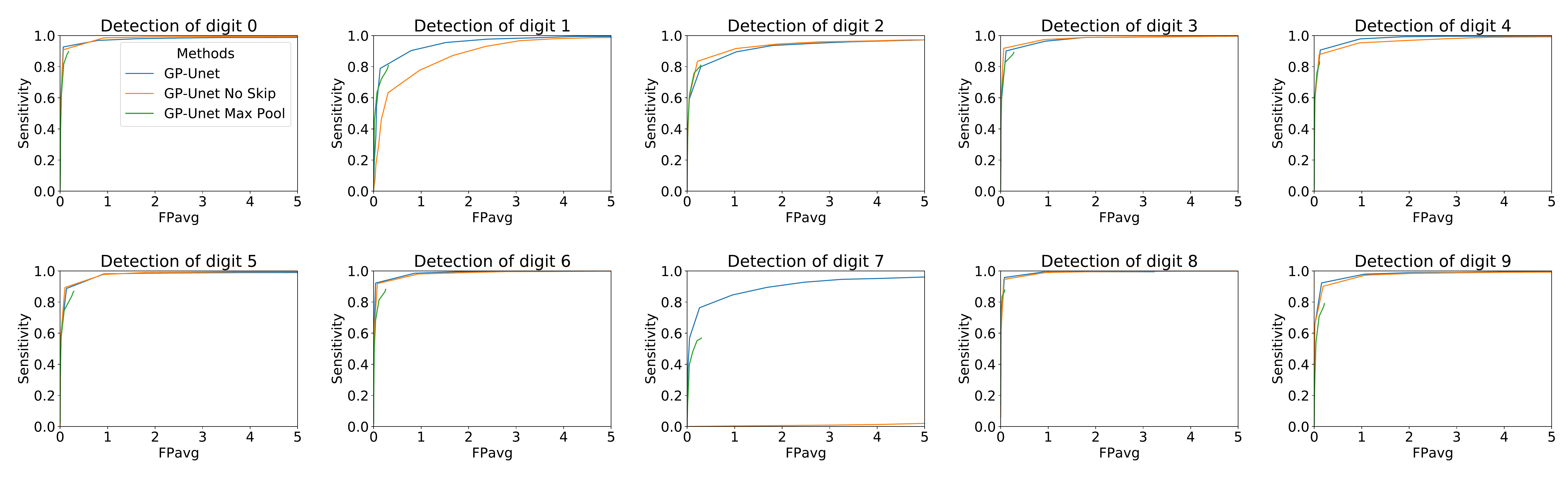}
\caption{\textbf{FROC MNIST architecture.} Each subplot corresponds to the detection of a different digit. GP-Unet is the standard GP-Unet architecture. In GP-Unet No Skip, blockwise skip connections are removed. In GP-Unet Max Pool, the global average pooling is replaced by global max pooling.}
\label{fig:attMapHip}
\end{figure}

\begin{table}[!htb]
\caption{\textbf{FAUCs MNIST architecture.} Each row corresponds to the detection of a different digit. 95 percent confidence interval is indicated in brackets. The average and standard deviation of the performance of each method across all digits is given in the last row. Best performance are indicated in bold.}
\centering
\begin{tabular}{cccc}
\hline
    & GP-Unet       & GP-Unet no skip & GP-Unet pax pooling \\ \hline
0   & \textbf{97.4 (96.3 - 98.3)} & \textbf{98.0 (97.5 - 98.5)}  & 89.2 (87.2 - 91.3)   \\
1   & \textbf{94.4 (93.6 - 95.2)} & 86.9 (85.6 - 88.2)  & 80.0 (77.2 - 82.6)   \\
2   & \textbf{91.7 (90.4 - 92.9)} & \textbf{92.9 (91.4 - 94.3)}  & 80.3 (77.5 - 83.0)   \\
3   & \textbf{97.3 (96.6 - 97.9)} & \textbf{97.9 (97.4 - 98.5)}  & 88.6 (86.4 - 90.8)   \\
4   & \textbf{97.8 (97.2 - 98.3)} & \textbf{96.2 (95.2 - 97.0)}  & 82.9 (80.2 - 85.5)   \\
5   & \textbf{97.1 (96.4 - 97.8)} & \textbf{97.6 (97.1 - 98.1)}  & 86.1 (83.6 - 88.6)   \\
6   & \textbf{98.6 (98.1 - 99.1)} & \textbf{98.1 (97.5 - 98.7)}  & 87.6 (85.2 - 89.9)   \\
7   & \textbf{89.3 (87.6 - 90.9)} & 0.8 (0.5 - 1.2)     & 56.3 (52.8 - 59.8)   \\
8   & \textbf{98.8 (98.2 - 99.2)} & \textbf{98.8 (98.4 - 99.1)}  & 87.7 (85.3 - 89.9)   \\
9   & \textbf{97.6 (96.8 - 98.2)} & \textbf{96.9 (96.1 - 97.6)}  & 78.4 (75.5 - 81.1)   \\
\hdashline[1.5pt/5pt]
Average & 96.0 +/- 3.0        & 86.4 +/- 28.7        & 81.7 +/- 9.3          \\ \hline
\end{tabular}
\end{table}

\begin{table}[!htb]
\caption{\textbf{Sensitivity MNIST architecture.} Each row corresponds to the detection of a different digit. 95 percent confidence interval is indicated in brackets. The average and standard deviation of the performance of each method across all digits is given in the last row. Best performance are indicated in bold.}
\centering
\begin{tabular}{cccc}
\hline
    & GP-Unet       & GP-Unet no skip & GP-Unet pax pooling \\ \hline
0   & \textbf{92.7 (91.4 - 93.9)} & \textbf{90.9 (89.6 - 92.3)}  & 81.8 (80.1 - 83.5)   \\
1   & \textbf{78.9 (77.3 - 80.4)} & 63.3 (61.5 - 65.0)  & 71.9 (70.1 - 73.8)   \\
2   & \textbf{80.0 (78.2 - 81.8)} & \textbf{83.4 (81.7 - 85.1)}  & 75.8 (74.0 - 77.7)   \\
3   & 90.1 (88.8 - 91.5) & \textbf{91.8 (90.7 - 93.0)}  & 82.8 (81.1 - 84.5)   \\
4   & \textbf{90.7 (89.3 - 92.1)} & 87.8 (86.3 - 89.2)  & 76.1 (74.1 - 78.1)   \\
5   & \textbf{88.7 (87.2 - 90.2)} & \textbf{89.4 (87.9 - 90.8)}  & 74.7 (72.7 - 76.8)   \\
6   & \textbf{92.2 (91.0 - 93.5)} & \textbf{91.9 (90.5 - 93.3)}  & 81.4 (79.5 - 83.4)   \\
7   & \textbf{76.3 (74.6 - 78.1)} & 0.6 (0.3 - 0.9)     & 55.2 (52.9 - 57.5)   \\
8   & \textbf{95.8 (95.0 - 96.5)} & \textbf{94.5 (93.7 - 95.4)}  & 83.7 (82.1 - 85.3)   \\
9   & \textbf{92.3 (91.1 - 93.5)} & 90.1 (88.8 - 91.5)  & 70.8 (68.8 - 72.9)   \\
\hdashline[1.5pt/5pt]
Average & 87.8 +/- 6.4        & 78.4 +/- 27.3        & 75.4 +/- 8.0          \\ \hline
\end{tabular}
\end{table}

\begin{table}[!htb]
\caption{\textbf{FPavg MNIST architecture.} Each row corresponds to the detection of a different digit. 95 percent confidence interval is indicated in brackets. The average and standard deviation of the performance of each method across all digits is given in the last row. Best performance are indicated in bold.}
\centering
\begin{tabular}{cccc}
\hline
    & GP-Unet       & GP-Unet no skip & GP-Unet pax pooling \\ \hline
0   & \textbf{0.07 (0.05 - 0.08)} & \textbf{0.07 (0.05 - 0.09)}     & \textbf{0.08 (0.06 - 0.10)}      \\
1   & \textbf{0.14 (0.11 - 0.16)} & 0.30 (0.26 - 0.35)     & \textbf{0.16 (0.13 - 0.19)}      \\
2   & 0.29 (0.25 - 0.32) & 0.22 (0.18 - 0.25)     & \textbf{0.15 (0.12 - 0.17)}      \\
3   & 0.11 (0.09 - 0.14) & \textbf{0.06 (0.05 - 0.08)}     & \textbf{0.09 (0.07 - 0.12)}      \\
4   & 0.13 (0.10 - 0.15) & 0.10 (0.08 - 0.13)     & \textbf{0.06 (0.04 - 0.07)}      \\
5   & 0.13 (0.11 - 0.16) & \textbf{0.10 (0.08 - 0.12)}     & \textbf{0.09 (0.06 - 0.11)}      \\
6   & \textbf{0.04 (0.03 - 0.06)} & 0.08 (0.06 - 0.10)     & 0.11 (0.09 - 0.14)      \\
7   & \textbf{0.26 (0.23 - 0.29)} & 1.99 (1.98 - 2.00)     & \textbf{0.21 (0.17 - 0.24)}      \\
8   & 0.08 (0.06 - 0.10) & 0.06 (0.05 - 0.08)     & \textbf{0.03 (0.02 - 0.04)}      \\
9   & 0.16 (0.13 - 0.18) & 0.19 (0.16 - 0.22)     & \textbf{0.10 (0.08 - 0.13)}      \\
\hdashline[1.5pt/5pt]
Average & 0.14 +/- 0.08      & 0.32 +/- 0.56          & 0.11 +/- 0.05           \\ \hline
\end{tabular}
\end{table}

\begin{table}[!htb]
\caption{\textbf{FNavg MNIST architecture.} Each row corresponds to the detection of a different digit. 95 percent confidence interval is indicated in brackets. The average and standard deviation of the performance of each method across all digits is given in the last row. Best performance are indicated in bold.}
\centering
\begin{tabular}{cccc}
\hline
    & GP-Unet       & GP-Unet no skip & GP-Unet pax pooling \\ \hline
0   & \textbf{0.11 (0.08 - 0.14)} & \textbf{0.13 (0.10 - 0.16)}     & 0.30 (0.25 - 0.35)      \\
1   & \textbf{0.35 (0.30 - 0.40)} & 0.60 (0.53 - 0.67)     & 0.49 (0.42 - 0.56)      \\
2   & 0.30 (0.25 - 0.34) & \textbf{0.25 (0.21 - 0.29)}     & 0.39 (0.34 - 0.45)      \\
3   & \textbf{0.15 (0.12 - 0.18)} & \textbf{0.13 (0.10 - 0.16)}     & 0.26 (0.22 - 0.31)      \\
4   & \textbf{0.12 (0.10 - 0.15)} & 0.18 (0.14 - 0.21)     & 0.37 (0.31 - 0.42)      \\
5   & \textbf{0.14 (0.11 - 0.16)} & \textbf{0.13 (0.11 - 0.16)}    & 0.36 (0.30 - 0.41)      \\
6   & \textbf{0.11 (0.08 - 0.13)} & \textbf{0.10 (0.08 - 0.12)}     & 0.26 (0.22 - 0.30)      \\
7   & \textbf{0.36 (0.31 - 0.41)} & 1.58 (1.42 - 1.74)     & 0.71 (0.62 - 0.80)      \\
8   & \textbf{0.07 (0.05 - 0.09)} & \textbf{0.09 (0.07 - 0.11)}     & 0.26 (0.21 - 0.31)      \\
9   & \textbf{0.11 (0.09 - 0.14)} & 0.15 (0.12 - 0.18)     & 0.47 (0.41 - 0.54)      \\
\hdashline[1.5pt/5pt]
Average & 0.18 +/- 0.10      & 0.33 +/- 0.44          & 0.39 +/- 0.13           \\ \hline
\end{tabular}
\end{table}

\begin{figure}[!htb]
\centering
\includegraphics[width=\textwidth]{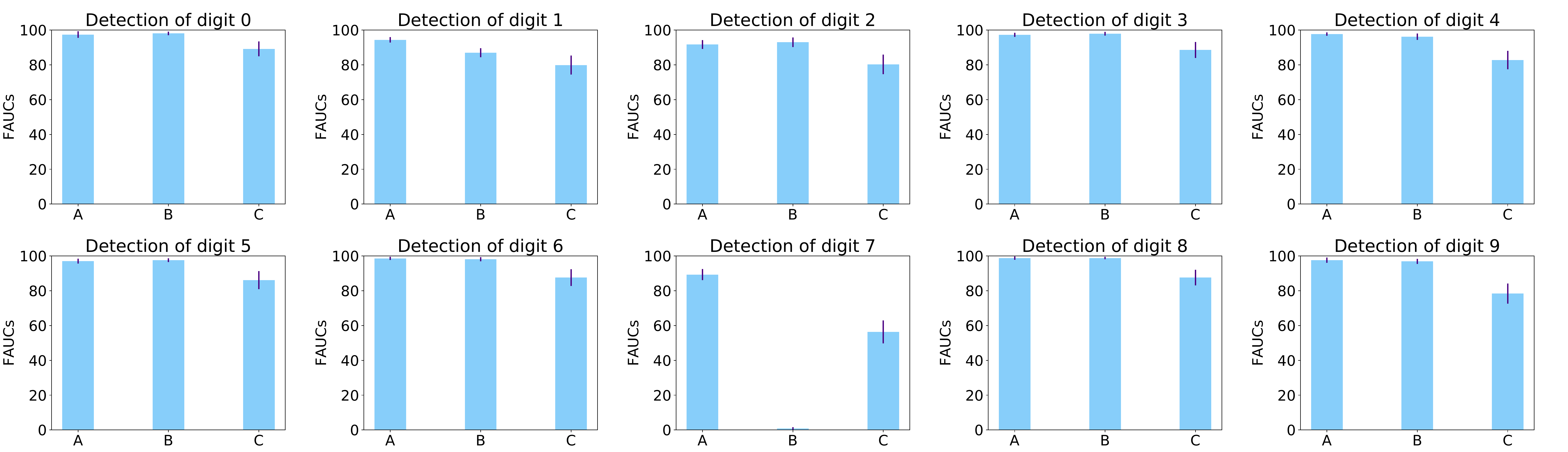}
\caption{\textbf{FAUCs on the MNIST dataset for different architectures of the proposed method: GP-Unet.} Each subplot corresponds to the detection of a different digit. FAUCs are displayed with confidence intervals computed by bootstrapping the test set. A: GP-Unet; B: GP-Unet without blockwise skip connections; C: GP-Unet with Global Max Pooling. For digit 7, GP-Unet without blockwise skip connections convergence to a very high value. We tried repeating the experiments with different random initializations of the weights and let the optimization run longer, but we achieved the same results. This supports the argument that adding blockwise skip connections eases the optimization.}
\label{fig:attMapHip}
\end{figure}

\begin{figure}[!htb]
\centering
\includegraphics[width=\textwidth]{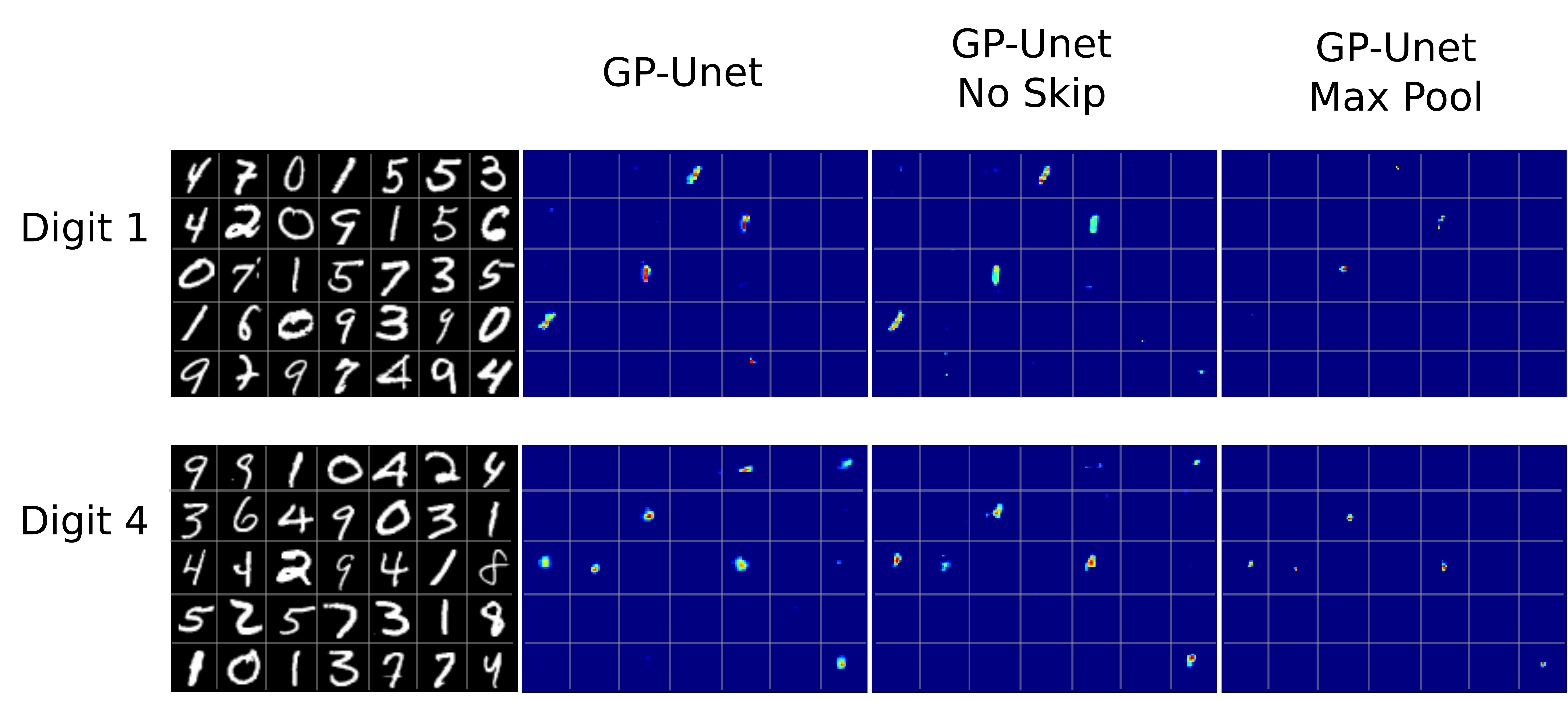}
\caption{\textbf{Attention Maps for the different architectures of GP-Unet.} In the first row, the digits 1 had to be detected. In the second row, digits 4 had to be detected. GP-Unet is the standard GP-Unet architecture. In GP-Unet No Skip, blockwise skip connections are removed. In GP-Unet Max Pool, the global average pooling is replaced by global max pooling.}
\label{fig:attMapHip}
\end{figure}

\end{document}